\documentclass[journal]{IEEEtran}
\usepackage{times}
\usepackage{epsfig}
\usepackage{graphicx}
\usepackage{amsmath}
\usepackage{amssymb}
\usepackage{colortbl}
\usepackage{multirow}
\usepackage{soul}
\usepackage{adjustbox}
\usepackage{array}
\usepackage{setspace}
\usepackage{lscape}
\usepackage{mathtools}

\usepackage{pgfplots}
\pgfplotsset{compat=newest}

\usepackage{enumitem}
\usepackage{array}
\usepackage{multirow}
\usepackage{rotating}
\usepackage{caption}
\usepackage{subcaption}
\usepackage{adjustbox}
\usepackage{cite}
\usepackage{amsmath,amssymb,amsfonts}
\usepackage{algorithmic}
\usepackage{graphicx}
\usepackage{textcomp}

\ifCLASSINFOpdf
\else
\fi
\begin{document}
%
\title{BioTouchPass2: Touchscreen Password Biometrics Using Time-Aligned Recurrent Neural Networks}
%
%
%

\author{Ruben Tolosana,
        Ruben Vera-Rodriguez, Julian Fierrez,~\IEEEmembership{Member,~IEEE}, Javier Ortega-Garcia,~\IEEEmembership{Fellow,~IEEE}\\

\thanks{The authors are with the Biometrics and Data Pattern Analytics - BiDA Lab, Escuela Politecnica Superior, Universidad Autonoma de Madrid, 28049 Madrid, Spain (e-mail: ruben.tolosana@uam.es; ruben.vera@uam.es; julian.fierrez@uam.es, javier.ortega@uam.es).}}
\maketitle

\begin{abstract}
Passwords are still used on a daily basis for all kind of applications. However, they are not secure enough by themselves in many cases. This work enhances password scenarios through two-factor authentication asking the users to draw each character of the password instead of typing them as usual. The main contributions of this study are as follows: \textit{i)} We present the novel MobileTouchDB public database, acquired in an unsupervised mobile scenario with no restrictions in terms of position, posture, and devices. This database contains more than 64K on-line character samples performed by 217 users, with 94 different smartphone models, and up to 6  acquisition sessions. \textit{ii)} We perform a complete analysis of the proposed approach considering both traditional authentication systems such as Dynamic Time Warping (DTW) and novel approaches based on Recurrent Neural Networks (RNNs). In addition, we present a novel approach named Time-Aligned Recurrent Neural Networks (TA-RNNs)\footnote{Spanish Patent Application (P202030060)}. This approach combines the potential of DTW and RNNs to train more robust systems against attacks. 

A complete analysis of the proposed approach is carried out using both MobileTouchDB\footnote{https://github.com/BiDAlab/MobileTouchDB} and e-BioDigitDB\footnote{https://github.com/BiDAlab/eBioDigitDB} databases. Our proposed TA-RNN system outperforms the state of the art, achieving a final 2.38\% Equal Error Rate, using just a 4-digit password and one training sample per character. These results encourage the deployment of our proposed approach in comparison with traditional typed-based password systems where the attack would have 100\% success rate under the same impostor scenario.
\end{abstract}

\begin{IEEEkeywords}
Biometrics, passwords, handwriting, touch biometrics, TA-RNNs, RNN, DTW, MobileTouchDB, e-BioDigitDB\end{IEEEkeywords}

\IEEEpeerreviewmaketitle

\section{Introduction}
\IEEEPARstart{M}{obile} devices have become an indispensable tool for most people nowadays~\cite{MobileAddictive}. The rapid and continuous deployment of mobile devices around the world has been motivated not only by the high technological evolution and new features incorporated but also to the new internet infrastructures like 5G that allows the communication and use of social media in real time, among many other factors. In this way, both public and private sectors are aware of the importance of mobile devices for the society and are trying to deploy their services through user-friendly mobile applications ensuring data protection and high security. 

Passwords are still the most common way to authenticate users nowadays. They can range from Personal Identification Numbers (PIN) that require users to memorise them to One-Time Passwords (OTP) where the security system is in charge of selecting and providing to the user a different password each time it is required, e.g., sending messages to personal mobile devices or special tokens. We use passwords on a daily basis for all kinds of applications. However, are passwords secure enough? Apparently not, at least by themselves. Recent news put in evidence this fact, e.g., in January 2019 a total of 21 million passwords from all parts of the world were released together with their corresponding emails addresses~\cite{passwords_stolen_new}. This important problem is related not only to data breaches, but also to many other attack scenarios, as it has been pointed out in different studies~\cite{Bonneau_2012, Galbally_password_TIFS_2017}. First, it is common to use passwords based on sequential digits (e.g., \textit{``1 2 3 4 5 6"}), personal information such as birth dates, or simply words such as \textit{``password"} or \textit{``qwerty"} that are very easy to guess~\cite{worst_passwords}. Second, passwords that are typed on mobile devices such as tablets or smartphones are susceptible to ``smudge attacks", i.e., the deposition of finger grease traces on the touchscreen can be used by the impostors to guess lock patterns or passwords~\cite{Aviv_2010}. Finally, password-based authentication is also vulnerable to ``shoulder surfing". This type of attack is produced when the impostor can observe directly or use external recording devices to collect the user information. This attack has attracted the attention of many researchers in recent years due to the increased deployment of handheld recording devices and public surveillance infrastructures~\cite{Shukla_2014, Yue_2014}. So, if we know that traditional passwords are not secure enough by themselves, but they continue to be present in our lives, how can we improve this authentication scenario?

\begin{figure*}[t]
\begin{center}
   \includegraphics[width=0.98\linewidth]{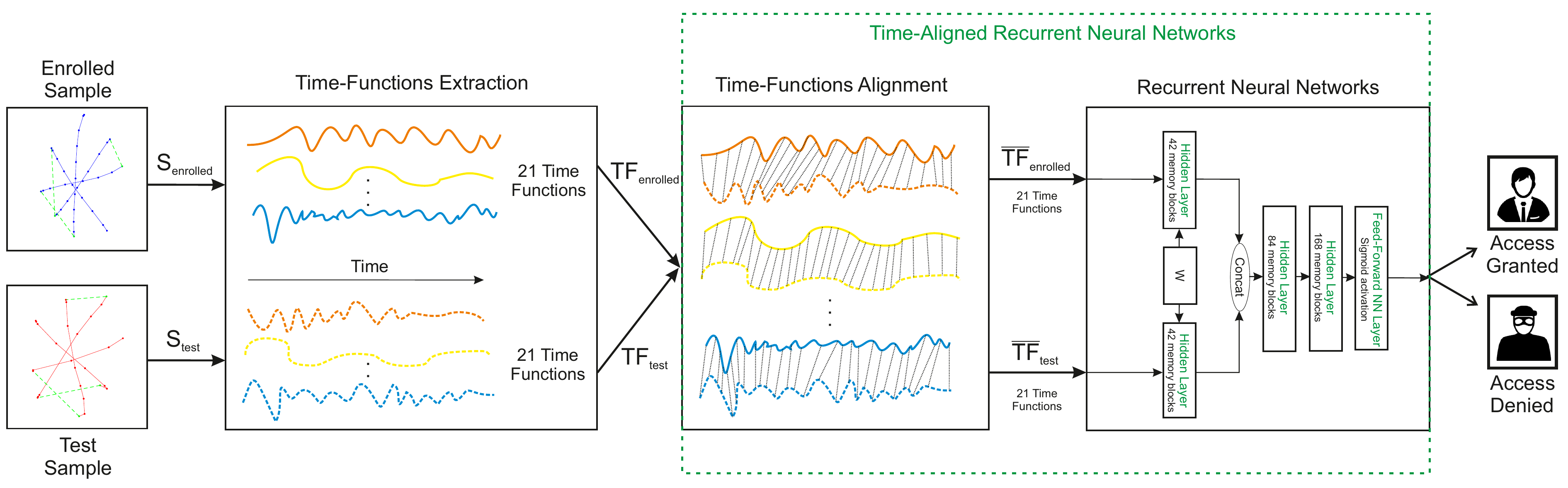}
\end{center}
   \caption{Architecture of our proposed touchscreen biometric system based on Time-Aligned Recurrent Neural Networks. $S$ denotes one character sample, and $TF$ and $\overline{TF}$ the original and pre-aligned 21 time functions, respectively. The Recurrent Neural Networks block is enlarged in Fig.~\ref{fig:LSTM_configuration_digits} for a better understanding. }
\label{fig:diagrama_TA-RNNs}
\end{figure*}

Two-factor authentication approaches have been very successful in the last years. These approaches are based on the combination of two authentication stages. For example: \textit{i)} the security platform sends the password to the personal email or mobile number of the claimed user, and \textit{ii)} the claimed user introduces this password for the final verification. Behavioral biometric information has also been considered for two-factor authentication approaches. In~\cite{Angulo_2011}, the authors proposed a two-factor verification system based on timing-related features for dynamic lock patterns, achieving a final average Equal Error Rate (EER) of 10.39\% for imitation attacks. A similar two-factor authentication approach was proposed in~\cite{Lacharme_2016} for traditional Android unlock patterns but considering biometric dynamic features related to the position of the finger, pressure, finger size, and accelerometer sensor achieving a final 15.0\% EER for imitation attacks. Two-factor authentication approaches have also been expanded to physiological biometric traits. In~\cite{Periocular_2017}, Jenkins \textit{et al.} proposed a system based on features extracted from periocular images acquired using an iPhone 5, achieving very good results for the task of identification. 


In this article we propose two-factor authentication approaches based on the incorporation of touch biometrics to password authentication systems, asking the users to draw each character of the password on the touchscreen instead of typing them as usual. One example of use that motivates our proposed approach is on internet payments with credit cards. Banks usually send a password (typically between 6 and 8 characters) to the user. This password must be inserted by the user in the security platform in order to complete the payment. Our proposed approach enhances such scenario by including a second authentication factor based on touch interaction biometrics.

The main contributions of this study are related to the novel MobileTouchDB database, our proposed architecture, and the competitive results obtained with respect to related research: 
\begin{itemize}

\item We present and describe the acquisition process of the new MobileTouchDB database. This database contains more than 64K on-line character samples performed by 217 users, using 94 different smartphone models. MobileTouchDB considers an unsupervised mobile scenario with a maximum of 6 captured sessions per subject and it is publicly available in GitHub\footnote{https://github.com/BiDAlab/MobileTouchDB}. 

\item The MobileTouchDB database opens the doors to many different applications: \textit{i)} analyse the discriminative power of novel human touch interaction dynamics, \textit{ii)} enhance traditional password authentication systems through the incorporation of touch biometric information as a second level of user authentication, and \textit{iii)} analyse the way we interact with mobile devices on a daily basis in order to enhance continuous authentication systems.

\item We present a novel approach named Time-Aligned Recurrent Neural Networks (TA-RNNs). Fig.~\ref{fig:diagrama_TA-RNNs} represents the general architecture of our proposed approach. It combines the potential of Dynamic Time Warping (DTW) and Recurrent Neural Networks (RNNs) to train more robust systems against attacks.

\item We perform a complete analysis of our proposed approach using both MobileTouchDB and e-BioDigitDB public databases. Three experiments are considered: \textit{i)} one-character analysis in order to evaluate the discriminative power of each character, \textit{ii)} character combination analysis so as to measure the robustness of our proposed approach when increasing the length of the passwords from 1 to 9 characters, and \textit{iii)} template update analysis. 

\item We compare our proposed TA-RNN system with both traditional and state-of-the-art authentication systems. Our proposed TA-RNN system outperforms the state of the art, achieving a final 2.38\% EER, using just a 4-digit password and one training sample per character.

\item We demonstrate the application of TA-RNNs for other time sequence tasks, i.e., on-line handwritten signature verification, outperforming in large margin the state of the art as well.

\end{itemize}

MobileTouchDB can be also useful for other research lines, e.g.: \textit{i)} user-dependent effects~\cite{yager2010biometric}, and development of user-dependent methods for handwriting recognition~\cite{2018_INFFUS_MCSreview2_Fierrez}, \textit{ii)} the neuromotor processes involved in writing over touchscreens~\cite{ferrer2018idelog, 2019_BookLogNormal_ComplexSignTouch_Vera}, \textit{iii)} sensing factors in obtaining representative and clean touch interaction signals~\cite{2015_IEEEAccess_InterSign_Tolosana, 2011_QualityBio_FAlonso}, \textit{iv)} human-device interaction factors involving touchscreen signals~\cite{harbach2016anatomy, 2018_TIFS_Swipe_Fierrez}, and development of improved interaction methods, and \textit{v)} population statistics around touch interaction signals, and development of new methods aimed at recognising or serving particular population groups~\cite{2018_IETB_DetectChildTouch_Acien}.

\begin{table*}[t]
\centering
\caption{Comparison of different touchscreen password biometric approaches for mobile scenarios.}
\label{table_related_works}
\scalebox{0.85}{
\begin{tabular}{cccccc}
\textbf{Study}             & \textbf{Classifiers}   & \multicolumn{3}{c}{\textbf{Verification Performance}}                                                                         & \textbf{\# Participants (Database)} \\
\multicolumn{1}{l}{}       & \multicolumn{1}{l}{}   & \textbf{Best Result}                                                  & \textbf{Password Length} & \textbf{Training Samples} & \multicolumn{1}{l}{}                \\ \hline
Kutzner \textit{et al.} (2015) \\ \cite{Travieso_2015}                    & Bayes-Nets, $K$Star, $k$NN & \begin{tabular}[c]{@{}c@{}}FAR = 10.42\%\\ FRR = unknown\end{tabular} & 8                  & 12                         & 32 (Unavailable DB)                 \\ \hline
Nguyen \textit{et al.} (2017) \\ \cite{Sae_Bae_2017}                     & DTW                    & EER = 4.84\%                                                          & 4                  & 5                          & 20 (Unavailable DB)                 \\ \hline
Tolosana \textit{et al.} (2019) \\ \cite{2019_TMC_BioTouchPass_Tolosana}               & DTW, RNNs              & EER = 3.80\%                                                          & 7                 & 3                          & 93 (e-BioDigitDB)                   \\ \hline
Le Lan \textit{et al.} (2019) \\ \cite{le2019securing}                  & RNNs                   & EER = 4.90\%                                                          & 4                  & 4                          & 93 (e-BioDigitDB)                   \\ \hline
Tolosana \textit{et al.} (2019) \\ \cite{2019_CVPRw_MobileTouch_Tolosana}              & DTW                    & EER = 5.90\%                                                          & 9                  & 1                          & 217 (MobileTouchDB)                 \\ \hline
\textbf{Proposed Approach} & \textbf{TA-RNNs}        & \textbf{EER = 2.38\%}                                                 & \textbf{4}        & \textbf{1}                 & \textbf{217 (MobileTouchDB)}        \\ \hline
\end{tabular}
}
\end{table*}

The remainder of the paper is organised as follows. Sec.~\ref{related_works} summarises related works in touchscreen password biometrics. Sec.~\ref{proposed_Approach} describes our proposed TA-RNN biometric system. Sec.~\ref{sec:databases} describes both e-BioDigitDB and the novel MobileTouchDB database considered in the experiments of this article. Sec.~\ref{experimental_protocol} and \ref{experimental_results} describe the experimental protocol and results achieved using our proposed approach, respectively. Finally, Sec.~\ref{conclusions} draws the final conclusions and points out some future work lines.

\section{Related Works}\label{related_works}
Touch biometrics are becoming a very attractive way to verify users on mobile devices~\cite{2018_TIFS_Swipe_Fierrez, eBioSign_journal}. Table~\ref{table_related_works} summarises the most relevant approaches in the area of touchscreen password biometrics on mobile scenarios. For each study, we include information related with classifiers and databases considered. We also report in Table~\ref{table_related_works} the best verification performance results in addition to other important system settings such as the length of the password, and the number of training samples. Regarding the attack scenario, all studies consider the case in which impostors know the password of the user, i.e., an \textit{imitation attack}~\cite{2018_HanbookBioAntiSpoofing_signature_Tolosana}. 


In~\cite{Travieso_2015}, Kutzner \textit{et al.} asked the users to perform an 8-digit password on the screen of a tablet device. For each handwritten password, a total of 25 static and dynamic features were extracted and tested using different authentication algorithms such as Bayes-Nets, $K$Star, and $k$NN. Their proposed approach achieved a final 10.42\% False Acceptance Rate (FAR). The False Rejection Rate (FRR) is not available. However, the authentication scenario considered in this preliminary approach restricted the deployment of the technology in real mobile applications as \textit{i)} the authors considered a large number of training samples (i.e., 12), and \textit{ii)} it seems to be only applicable to devices with large screens (such as tablets) as it would be very difficult for the users to perform such a long password (8 characters) on a screen of much smaller size. In~\cite{Sae_Bae_2017}, Nguyen \textit{et al.} evaluated the use of handwritten touch biometrics for PIN-based authentication systems. Their proposed authentication approach overcame some of the drawbacks previously cited as they asked users to draw each character of a 4-digit PIN one by one. A final 4.84\% EER was achieved using 5 training samples, and a biometric system composed of 5 dynamic features and DTW algorithm. In~\cite{2019_TMC_BioTouchPass_Tolosana}, we released e-BioDigitDB, the first touchscreen password biometric public database, containing only numbers from 0 to 9. In addition, we reported a benchmark evaluation of biometric authentication using two different state-of-the-art approaches: \textit{i)} DTW, and \textit{ii)} RNNs. Despite the good results obtained using RNNs (outperforming DTW for 50\% of the numerical characters), DTW still improved the RNN results with an average 0.5\% EER absolute improvement. In~\cite{le2019securing}, Le Lan \textit{et al.} also evaluated RNN approaches using the e-BioDigitDB database. Their proposed approach achieved a final 4.90\% EER using 4-digit passwords and 4 training samples per digit. Finally, a preliminary study of the work presented here was published in~\cite{2019_CVPRw_MobileTouch_Tolosana}. In that work we performed an initial analysis of MobileTouchDB considering only a DTW baseline system so as to provide an easily reproducible framework.

The study presented here extends the preliminary analysis carried out in~\cite{2019_CVPRw_MobileTouch_Tolosana}. The main improvements over~\cite{2019_CVPRw_MobileTouch_Tolosana} are:

\begin{itemize}
\item Sec.~\ref{related_works} has been included to survey and compare advantages and limitations of recent research on touchscreen password biometrics on mobile scenarios.

\item Our preliminary touch biometric system in~\cite{2019_CVPRw_MobileTouch_Tolosana} based on DTW has been extended by incorporating RNNs and presenting a novel approach not published yet, named TA-RNNs (Sec.~\ref{proposed_Approach}). This approach combines the potential of DTW and RNNs with a Siamese architecture to train more robust systems against attacks.

\item The results achieved in the present study outperform the results achieved in~\cite{2019_CVPRw_MobileTouch_Tolosana}, and also the state of the art, achieving a final 2.38\% EER. This improvement is achieved considering more user-friendly scenarios, i.e., using just a 4-digit password and one training sample per character.
\end{itemize}

\section{Proposed Approach}\label{proposed_Approach}
This section describes our proposed Time-Aligned Recurrent Neural Networks for touchscreen password biometrics. A graphical representation is included in Fig.~\ref{fig:diagrama_TA-RNNs}.

\subsection{Time-Functions Extraction}\label{subsec_FeatureExtractor}
Our proposed touchscreen password biometric system is based on time functions. For each character sample acquired (i.e., $S_{enrolled}$ and $S_{test}$ in Fig.~\ref{fig:diagrama_TA-RNNs}), signals related to $X$ and $Y$ spatial coordinates are used to extract a set of 21 time functions (i.e., $TF_{enrolled}$ and $TF_{test}$ in Fig.~\ref{fig:diagrama_TA-RNNs}), similar to~\cite{2019_TMC_BioTouchPass_Tolosana}. This set of 21 time functions is widely used in on-line signature verification~\cite{2018_IEEEAccess_RNN_Tolosana,martinez14mobileSignRobustPerf}. The complete set of time functions is described in Table~\ref{tabla:tablaLocalFeatures}.

\begin{table}[tb]
\centering
\caption{Set of time functions considered in this work.}
\begin{adjustbox}{width=0.40\textwidth}
\begin{tabular}{p{1cm}|| p{7cm}}
\hline
\# & Feature \\
\hline \hline
1 & \textit{X}-coordinate: $x_n$  \\
\hline
2 & \textit{Y}-coordinate: $y_n$  \\
\hline
3 & Path-tangent angle: $\theta_n$  \\
\hline
4 & Path velocity magnitude: $v_n$ \\
\hline
5 & Log curvature radius: $\rho_n$ \\
\hline
6 & Total acceleration magnitude: $a_n$ \\
\hline
7-12 & First-order derivative of features 1-6: $\dot{x_n},\dot{y_n},\dot{\theta_n},\dot{v_n},\dot{\rho_n},\dot{a_n}$ \\
\hline
13-14 & Second-order derivative of features 1-2: $\ddot{x_n},\ddot{y_n}$ \\
\hline
15 & Ratio of the minimum over the maximum speed over a 5-samples window: $v^r_n$ \\
\hline
16-17 & Angle of consecutive samples and first-order derivative: $\alpha_n$, $\dot{\alpha_n}$ \\
\hline
18 & Sine of the angle of consecutive samples: $s_n$ \\
\hline
19 & Cosine of the angle of consecutive samples: $c_n$ \\
\hline
20 & Stroke length to width ratio over a 5-samples window: $r^5_n$ \\
\hline
21 & Stroke length to width ratio over a 7-samples window: $r^7_n$ \\
\hline
\end{tabular}
\end{adjustbox}
\label{tabla:tablaLocalFeatures}
\end{table}

\subsection{Time-Functions Alignment}\label{subsec_SWDTW}
One crucial point when comparing the similarity among time sequences is the proper alignment of them prior to calculating the similarity score through distance measurement functions (e.g., the Euclidean distance). DTW is one of the most popular algorithms in the literature, in particular for signature biometrics~\cite{Marcos_matching}. The goal of DTW is to find the optimal warping alignment path of a pair of time sequences $A$ and $B$ that minimises a given distance measure. The algorithm can be defined as follows. Let's define two sequences as:

\begin{equation}
\begin{array}{l}
 A =  a_1 ,a_2 ,...,a_n ,...,a_N  \\
 B = b_1 ,b_2 ,...,b_m ,...,b_M  \\
 \end{array}
\end{equation}

\noindent and a distance measure among sequence samples as:

\begin{equation}
d(a_n,b_m) = (a_n  - b_m)^2  \label{eq:distance}
\end{equation}

\noindent A warping path can be defined as:

\begin{equation}
C = c_1 ,c_2 ,...,c_k
,...,c_K
\end{equation}

\noindent where each $c_k$ represents a correspondence
$(n,m)$ between samples of $A$ and $B$. The
initial condition of the algorithm is set to:

\begin{equation}
g_1  = g(1,1) = d(a_1,b_1) \cdot w(1)
\end{equation}

\noindent where $g_k$ represents the accumulated distance after $k$
steps and $w(k)$ is a weighting factor that must be defined. For
each iteration, $g_k$ is computed as:

\begin{equation}
g_k  = g(n,m) = \mathop {\min }\limits_{c_{k - 1}} \left[ {g_{k - 1}
+ d(c_k) \cdot w(k)} \right] \label{eq:gk}
\end{equation}

\noindent until the $N$'th and $M$'th sample of both sequences
respectively is reached. The resulting normalised distance is:

\begin{equation}
D(A,B) = \frac{{g_K }}{{\sum\nolimits_{k=1}^K
{w(k)} }}
\end{equation}

\noindent where $\sum {w(k)} $ compensates the effect of the length of the sequences. The weighting factors $w(k)$ are defined in order to restrict which correspondences among samples of both sequences are allowed. In this work, only three transitions with the same value equal to 1 are allowed for the computation of $g_k$, which is the most common implementation found in the literature. Consequently, Eq.~(\ref{eq:gk}) becomes:

\begin{equation}
g_k  = g(n,m) = \min \left[ \begin{array}{l}
 g(n,m - 1) + d(a_n,b_m) \\
 g(n - 1,m - 1) + d(a_n,b_m) \\
 g(n - 1,m) + d(a_n,b_m) \\
 \end{array} \right]
\end{equation}

In this study we consider an improved version of the basic DTW algorithm named Sliding Window Dynamic Time Warping (SW-DTW), proposed in~\cite{folgado2018time}. This updated version overcomes some of the incorrect alignments generated by the original DTW. This is produced due to the original DTW considers an element-to-element distance. SW-DTW modifies the original distance in Eq.~(\ref{eq:distance}) to consider the context by incorporating a weighted average of the neighbouring distances.

%
%
%
%
%

In our proposed approach, SW-DTW is applied in a first stage in order to convert the 21 original time functions (i.e., $TF_{enrolled}$ and $TF_{test}$ in Fig.~\ref{fig:diagrama_TA-RNNs}) into 21 pre-aligned time functions (i.e., $\overline{TF}_{enrolled}$ and $\overline{TF}_{test}$ in Fig.~\ref{fig:diagrama_TA-RNNs}) before introducing them to the RNNs. This way our proposed RNN system is able to extract more meaningful features as time sequences have been previously normalised through the optimal warping path.

%

\subsection{Recurrent Neural Networks}\label{subsec_SRNNs}
New trends based on the use of RNNs, which is a specific neural network architecture, are becoming more and more important nowadays for modelling sequential data with arbitrary length. In this study we adapt the original verification system proposed in~\cite{2018_IEEEAccess_RNN_Tolosana} for on-line handwritten signature to touchscreen password biometrics. In~\cite{2018_IEEEAccess_RNN_Tolosana} we proposed RNN systems based on a Siamese architecture. The main goal was to learn a dissimilarity metric from data minimising a discriminative cost function that drives the dissimilarity metric to be small for pairs of genuine samples from the same subject, and higher for pairs of samples coming from different subjects. In particular, we consider in this study a Bidirectional Long Short-Term Memory (BLSTM) network, which allows access to past, present, and future context, achieving much better results compared with the original LSTM.

\begin{figure}[t]
\begin{center}
   \includegraphics[width=\linewidth]{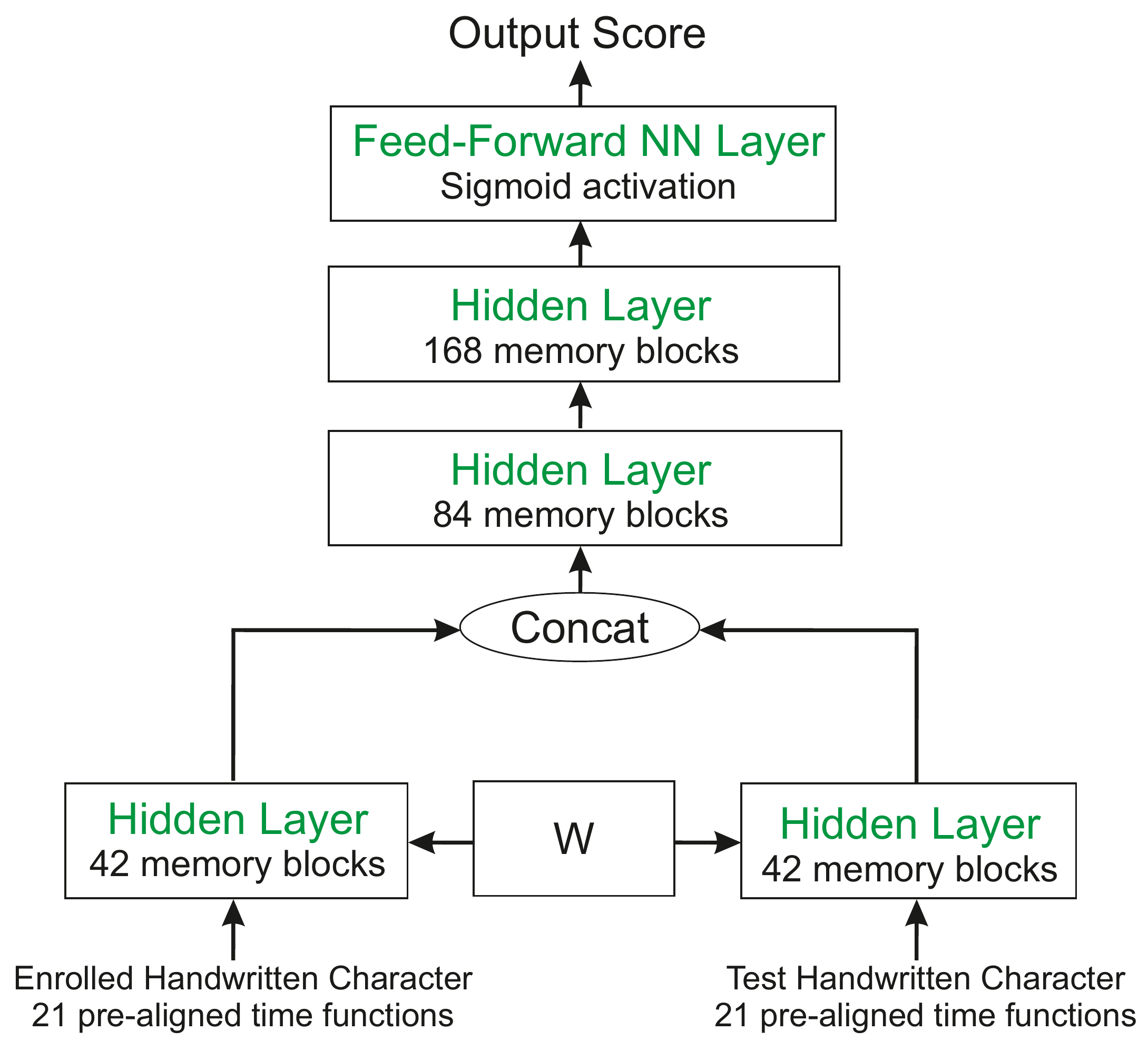}
\end{center}
   \caption{Proposed writer-independent BLSTM Siamese architecture.}
\label{fig:LSTM_configuration_digits}
\end{figure}

Fig.~\ref{fig:LSTM_configuration_digits} shows our proposed writer-independent BLSTM touch biometric system based on a Siamese architecture. For the input of the system, we feed the network with as much information as possible, i.e., all 21 pre-aligned time functions per character previously normalised through SW-DTW. The first layer is composed of two BLSTM hidden layers with 42 memory blocks each, sharing the weights between them. The outputs of the first two parallel BLSTM hidden layers are concatenated and serve as input to the second layer, which corresponds to a BLSTM hidden layer with 84 memory blocks. Then, we add a third BLSTM hidden layer with 168 memory blocks. The activation functions considered in all BLSTM memory blocks are based on the standard approach~\cite{2018_IEEEAccess_RNN_Tolosana}. Finally, a feed-forward neural network layer with a sigmoid activation is considered, providing an output score for each pair of characters. It is important to highlight that our approach is trained to distinguish between genuine and impostor patterns from all characters and users. Thus, we just train one writer-independent system for all characters and users through a development dataset.

\section{Databases}\label{sec:databases}

\subsection{e-BioDigitDB}\label{subsec_eBioDigitDB}
e-BioDigitDB database is composed of 93 users and was originally captured in order to perform a preliminary study of handwritten passwords for touchscreen biometrics~\cite{2019_TMC_BioTouchPass_Tolosana}. This database comprises on-line handwritten numerical characters from 0 to 9 acquired using a Samsung Galaxy Note 10.1 general purpose tablet. This device has a 10.1-inch LCD display with a resolution of 1280$\times$800 pixels.

Regarding the acquisition protocol, subjects had to perform handwritten numerical characters one at a time, in a supervised and office-like scenario. Samples were collected in two sessions with a time gap of at least three weeks between them in order to consider inter-session variability, very important for behavioral biometric traits~\cite{2019_IETBiometrics_Aging_Tolosana}. For each session, users had to perform a total of 4 numerical sequences from 0 to 9 using the finger as input. Therefore, there are a total of 8 samples per numerical digit and user. 

\begin{figure}[t]
\begin{center}
   \includegraphics[width=\linewidth]{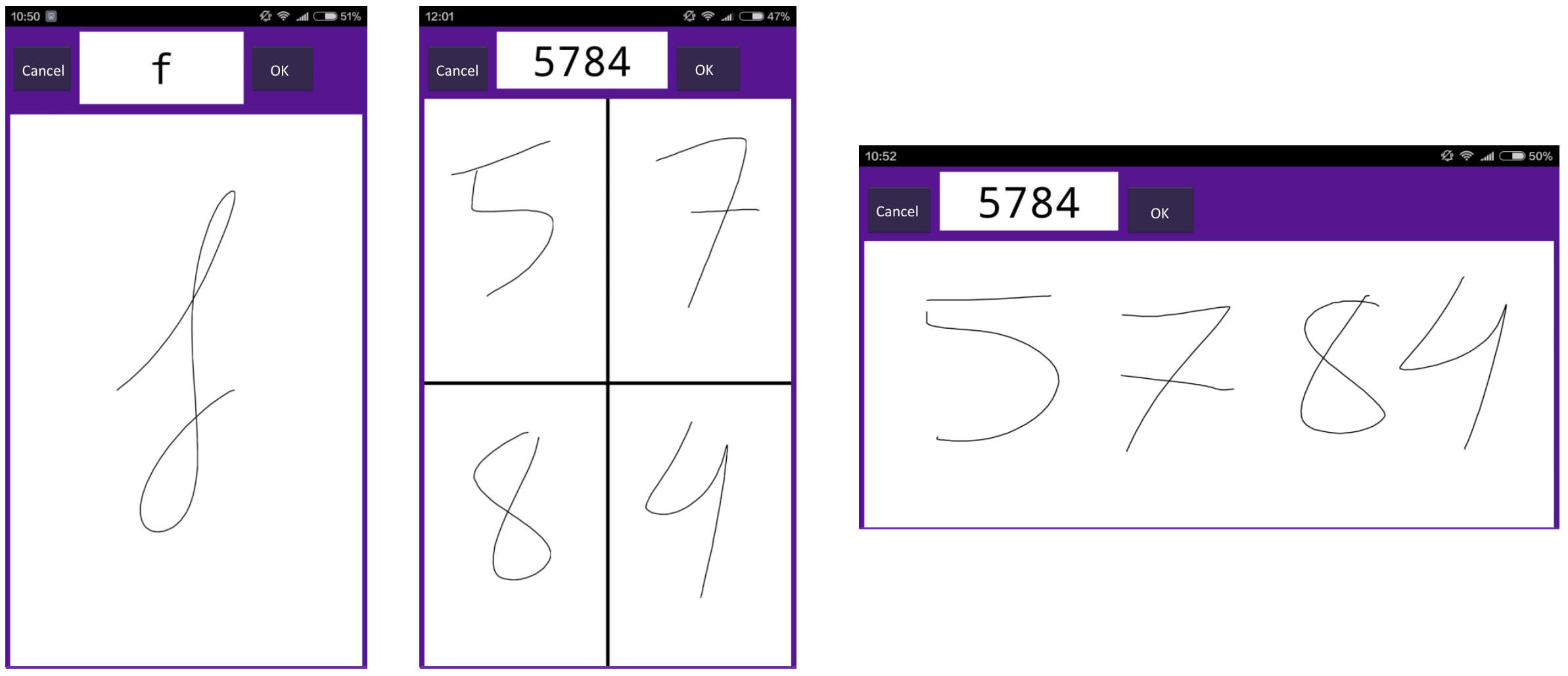}
\end{center}
   \caption{Different interfaces designed for the acquisition app. Both portrait and landscape orientations are considered in order to analyse different user experiences while drawing.}
\label{fig_examples_interface}
\end{figure}

\begin{figure*}[t]
\begin{center}
   \includegraphics[width=0.9\linewidth]{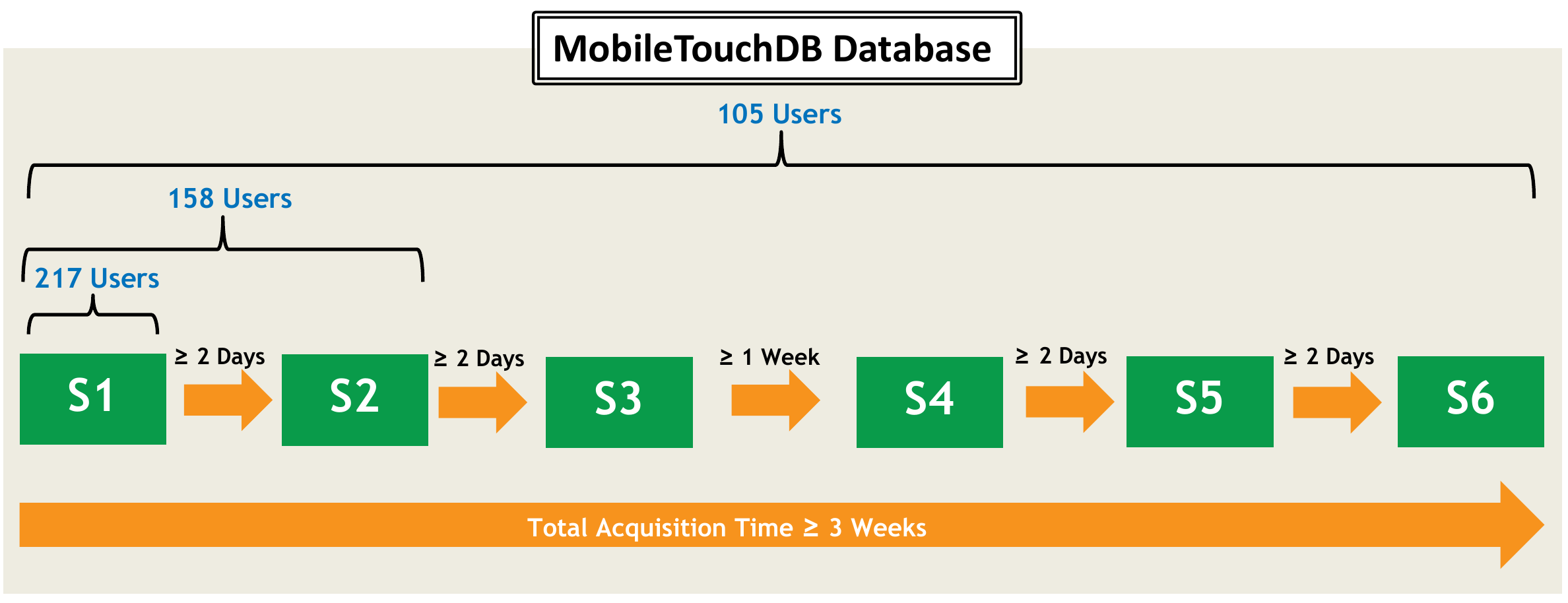}
\end{center}
   \caption{Description of the design and number of available users of the new MobileTouchDB database.}
\label{fig_experimental_protocol}
\end{figure*}


\subsection{MobileTouchDB}\label{subsec_MobileTouchDB}
MobileTouchDB is a novel handwritten character mobile touch biometric database composed of more than 64K on-line character samples performed by 217 users. For the acquisition, we implemented an Android application. Fig.~\ref{fig_examples_interface} represents the different interfaces designed for the acquisition. All interfaces are composed of: \textit{i)} the character/password to draw (top, middle) and two buttons ``OK'' (top, right) and ``Cancel'' (top, left) to press after drawing if the sample was good or bad respectively. If the sample was not good, then it was repeated; and \textit{ii)} a rectangular area to perform the character or password. In order to study an unsupervised mobile scenario, the acquisition app was uploaded to the Play Store. This way all participants could download and use the app on their own devices without any kind of supervision, simulating a practical scenario in which users can generate touchscreen information in any possible scenario, e.g., standing, sitting, walking, indoors, outdoors, etc. As a result, 94 different models from the following 16 brands were collected during the acquisition: \textit{Alcatel, Blackberry, BQ, Coolpad, Doogee, Google, Huawei, LeTV, LG, Motorola, OnePlus, Samsung, Sony, UMIDIGI, Xiaomi, and ZTE}. The acquisition app was designed to capture the following time signals: \textit{X} and \textit{Y} spatial coordinates, the area covered by the finger, timestamp, accelerometer, and gyroscope. However, information related to the area covered by the finger, accelerometer, and gyroscope was not available in some cases depending on how old the acquisition device was.

\begin{figure*}[!]
\begin{center}
   \includegraphics[width=0.70\linewidth]{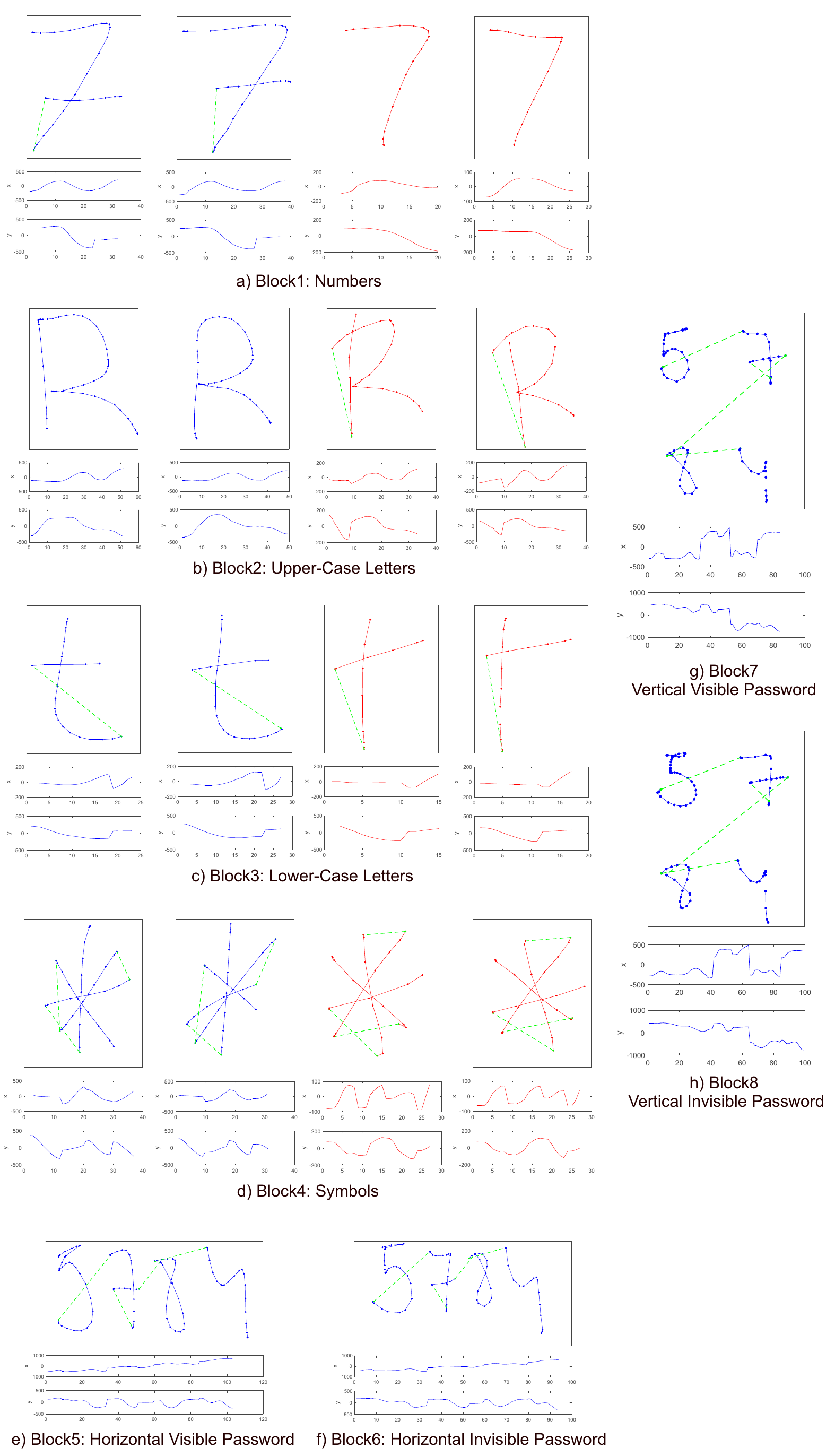}
\end{center}
   \caption{Example of the data collected in MobileTouchDB database. Blue and red colours represent samples drawn by different users. The green dashed lines indicate pen up trajectories between strokes. Curves under each character represent \textit{X} and \textit{Y} trajectories over time.}
\label{fig_examples_database}
\end{figure*}

The acquisition protocol considered in MobileTouchDB is depicted in Fig.~\ref{fig_experimental_protocol}. It comprises a total of 6 sessions (i.e., S1-S6) with different time gaps among them. It is important to highlight that in all sessions, the time gap refers to the minimum time between one user finishes a session and the following session is available. However, participants usually performed their corresponding sessions later on thanks to notifications sent automatically by the acquisition app to the users. Regarding the data acquired, each session comprises 8 different capturing blocks (i.e., from Block1 to Block8). Fig.~\ref{fig_examples_database} shows some examples of each of the eight acquisition blocks for two different users (indicated in blue and red colours). The green dashed lines indicate pen up trajectories between strokes. In Block1, we asked users to draw all numbers (from 0 to 9). Block2 and Block3 comprise upper- and lower-case letters respectively, with a total of 27 letters each. Block4 is composed of 8 different symbols (i.e., \textit{``?'', ``\#'', ``*'', ``@'', ``\%'', ``='', ``$\epsilon$'', and ``$\alpha$''}). It is important to remark that inside each block, characters were randomised before asking users to draw them. This way, each user performs a different character sequence in each session. From Block1 to Block4, the acquisition interface was designed as portrait to provide a better user experience (see Fig.~\ref{fig_examples_interface}, left). After finishing the first 4 blocks focused on performing one single character at a time (one sample per character), we asked users to draw passwords composed of 4 numbers (always \textit{``5 7 8 4''}) in different ways (6 samples in total). In Block5, users performed the password twice using a landscape orientation interface (see Fig.~\ref{fig_examples_interface}, right). We provided the users with a graphical visualization of the numbers while drawing them (i.e., visible mode). Then, in Block6, users had to repeat once the same task considered in Block5 but this time in an invisible mode, i.e., we did not provide to the users any visualization of the numbers while drawing them. The main motivation of this novel acquisition scenario is to protect us against shoulder surfing attacks, as commented in~\cite{Sae_Bae_2017}. In Block 7, users had to draw each number of the password inside each of the four available boxes (two times), considering first a visible mode (see Fig.~\ref{fig_examples_interface}, middle). Finally, in Block8 users had to repeat once the same task considered in Block7 but this time in an invisible mode. In both Block7 and Block8 the acquisition interface was kept portrait to analyse the user experience in different settings. 

Regarding the MobileTouchDB population statistics, 217 users completed the S1 acquisition session. S1 and S2 were completed by 159 users. Finally, a total of 105 users completed the six acquisition sessions. This participant reduction between S1 and S6 sessions is produced due to the challenging acquisition scenario considered in this study as it was completely unsupervised and comprised several acquisition sessions along time. Regarding the age distribution, 36.2\% of the participants are younger than 22 years old, 31.9\% are between 22 and 27 years old, and the remaining 31.9\% are older than 27 years old. Regarding the gender, 63\% of the participants were males, and 37\% females. 96\% of the population was righthanded.

\begin{table*}[h]
\centering
\caption{Performance as EER(\%) of each individual upper- and lower-case letter on the evaluation dataset of  MobileTouchDB.}
\label{exp1_mobileTouchDB_letters}
\begin{tabular}{c|cccccc|cccc}
     & DTW            & SW-DTW & RNNs          & TA-RNNs       &  &      & DTW            & SW-DTW         & RNNs          & TA-RNNs       \\ \cline{1-5} \cline{7-11} 
A    & 30.95          & 35,71  & 28,57          & \textbf{23,81} &  & a    & 30,95          & 28,57          & \textbf{16,67} & 19,05          \\
B    & 23.81          & 23,81  & 16,67          & \textbf{11,90} &  & b    & 23,81          & 19,05          & 19,05          & \textbf{16,67} \\
C    & 35.71          & 33,33  & \textbf{23,81} & \textbf{23,81} &  & c    & 40,48          & 38,10          & 28,57          & \textbf{26,19} \\
D    & 33.33          & 23,81  & 23,81          & \textbf{21,43} &  & d    & 19,05          & 19,05          & 19,05          & \textbf{14,29} \\
E    & 28,57          & 23,81  & 26,19          & \textbf{21,43} &  & e    & 35,71          & 30,95          & 26,19          & \textbf{21,43} \\
F    & 23,81          & 28,57  & 30,95          & \textbf{19,05} &  & f    & \textbf{19,05} & \textbf{19,05} & \textbf{19,05} & \textbf{19,05} \\
G    & 26,19          & 25,89  & 21,43          & \textbf{16,67} &  & g    & 21,43          & 23,81          & 19,05          & \textbf{16,67} \\
H    & 26,19          & 21,43  & 19,05          & \textbf{14,29} &  & h    & 28,57          & 30,95          & 23,81          & \textbf{16,67} \\
I    & \textbf{23,81} & 26,19  & 26,19          & \textbf{23,81} &  & i    & 30,95          & 33,33          & \textbf{19,05} & 21,43          \\
J    & 21,43          & 19,05  & 23,81          & \textbf{16,67} &  & j    & 26,19          & 23,81          & \textbf{19,05} & 21,43          \\
K    & 28,57          & 26,19  & 23,81          & \textbf{16,67} &  & k    & 16,67          & 16,67          & 16,67          & \textbf{14,29} \\
L    & 30,95          & 30,95  & \textbf{23,81} & 28,57          &  & l    & 28,57          & 28,57          & 19,05          & \textbf{16,67} \\
M    & 26,19          & 23,81  & 23,81          & \textbf{19,05} &  & m    & 28,57          & 21,43          & 21,43          & \textbf{16,67} \\
N    & 33,33          & 30,95  & 26,19          & \textbf{23,81} &  & n    & 28,57          & 23,81          & 19,05          & \textbf{16,67} \\
\~{N}    & 30,95          & 28,57  & 21,43          & \textbf{19,05} &  & \~{n}    & 23,81          & 23,81          & \textbf{16,67} & \textbf{16,67} \\
O    & 26,19          & 26,19  & \textbf{21,43} & \textbf{21,43} &  & o    & 26,19          & \textbf{23,81} & 26,19          & 26,19          \\
P    & 28,57          & 28,57  & \textbf{21,43} & 23,81          &  & p    & 33,33          & 33,33          & \textbf{21,43} & \textbf{21,43} \\
Q    & 30,95          & 26,19  & \textbf{23,81} & \textbf{23,81} &  & q    & 23,81          & 19,05          & \textbf{16,67} & \textbf{16,67} \\
R    & 23,81          & 21,43  & 21,43          & \textbf{16,67} &  & r    & 19,05          & 19,05          & 16,67          & \textbf{11,90} \\
S    & 28,57          & 28,57  & \textbf{21,43} & \textbf{21,43} &  & s    & 35,71          & 33,33          & 23,81          & \textbf{21,43} \\
T    & 23,81          & 21,43  & \textbf{16,67} & 21,43          &  & t    & 23,81          & 21,43          & 21,43          & \textbf{19,05} \\
U    & 35,71          & 35,71  & 26,19          & \textbf{23,81} &  & u    & 28,57          & 26,19          & \textbf{19,05} & \textbf{19,05} \\
V    & 38,10          & 35,71  & 33,33          & \textbf{30,95} &  & v    & 23,81          & 16,67          & \textbf{16,67} & 19,05          \\
W    & 26,19          & 21,43  & 19,05          & \textbf{14,29} &  & w    & 28,57          & 26,19          & \textbf{19,05} & \textbf{19,05} \\
X    & \textbf{19,05} & 21,43  & 21,43          & 21,43          &  & x    & 21,43          & \textbf{19,05} & \textbf{19,05} & \textbf{19,05} \\
Y    & 33,33          & 30,95  & 26,19          & \textbf{23,81} &  & y    & \textbf{19,05} & \textbf{19,05} & 21,43          & \textbf{19,05} \\
Z    & 33,33          & 30,95  & 28,57          & \textbf{23,81} &  & z    & 33,33          & 33,33          & 21,43          & \textbf{19,05} \\ \cline{1-5} \cline{7-11} 
Average & 28.57          & 27.06  & 23.72          & \textbf{20.99} &  & Average & 26.63          & 24.87          & 20.20          & \textbf{18.70}
\end{tabular}
\end{table*}

\section{Experimental Protocol}\label{experimental_protocol}
Three different experiments are considered: \textit{i)} one-character analysis in order to evaluate the discriminative power of each character, \textit{ii)} character combination analysis so as to measure the robustness of our proposed approach when increasing the length of the passwords from 1 to 9 characters, and \textit{iii)} analysis of the system performance when samples performed in different acquisition sessions are considered for training. Due to the large amount of information acquired in MobileTouchDB, in this paper we focus on characters performed one at a time. Complete passwords acquired from Block5 to Block8 will be analysed in future studies.

Both e-BioDigitDB and MobileTouchDB are divided into development and evaluation datasets. For the e-BioDigitDB, we consider the same experimental protocol proposed in~\cite{2019_TMC_BioTouchPass_Tolosana}. Thus, the first 50 users are considered for development whereas the remaining 43 users are left for the final evaluation of the systems. For MobileTouchDB, users are divided into development (175, $\approx$80\%) and evaluation (42, $\approx$20\%) datasets. Finally, the development dataset of both databases is also split into training ($\approx$80\%) and validation ($\approx$20\%) datasets in order to train the weights of the neural networks and select the best models. It is important to remark that for MobileTouchDB, all users included in the evaluation dataset (i.e., 42) contain the maximum number of acquisition sessions, i.e., 6. This way we can perform a fair evaluation of our trained models on different experimental conditions.

For the development of our proposed handwritten touch biometric systems (Sec.~\ref{subsec_exp1}), $Z$ genuine samples per character from the first session (i.e., 1 for MobileTouchDB and up to 4 for e-BioDigitDB) are used as enrolment samples, whereas the remaining genuine samples from the second session are used for testing. This way we consider the inter-session variability problem as genuine samples from different acquisition sessions are used as enrolment and testing samples, respectively. Finally, impostor scores are obtained by comparing the $Z$ enrolment samples with one genuine sample of each of the remaining users of the same database (simulating this way the imitation attack in which the impostor knows the password). 

For the final evaluation of our proposed touch biometric systems, different scenarios are generally considered regarding the number of available enrolment samples per user (i.e., $Z$vs1), in which the final score is performed as the average score of $Z$ one-to-one comparisons. In addition, for the character combination analysis (Sec.~\ref{subsec_exp2}), the final score is produced by fusing the different one by one character score comparisons using the sum of the scores~\cite{2018_INFFUS_MCSreview2_Fierrez}.

\section{Experimental Results}\label{experimental_results}

\begin{table}[h]
\centering
\caption{Performance as EER(\%) of each individual number and symbol on the evaluation dataset of MobileTouchDB.}
\label{exp1_mobileTouchDB_numbers}
\begin{tabular}{c|cccc}
     & DTW            & SW-DTW         & RNNs          & TA-RNNs       \\ \hline
0    & 28,57          & 23,81          & \textbf{21,43} & \textbf{21,43} \\
1    & 28,57          & 30,95          & \textbf{21,43} & \textbf{21,43} \\
2    & 28,57          & 30,95          & 23,81          & \textbf{16,67} \\
3    & 28,57          & 28,57          & 28,57          & \textbf{21,43} \\
4    & 28,57          & 30,95          & 23,81          & \textbf{21,43} \\
5    & 28,57          & 30,95          & 26,19          & \textbf{19,05} \\
6    & 30,95          & 28,57          & 28,57          & \textbf{23,81} \\
7    & 28,57          & 28,57          & 23,81          & \textbf{19,05} \\
8    & 23,81          & 21,43          & 19,05          & \textbf{14,29} \\
9    & 28,57          & 30,95          & 26,19          & \textbf{23,81} \\ \hline
Average & 28.33          & 28.57          & 24.29          & \textbf{20.24} \\
     &                &                & \textbf{}      &                \\
?    & \textbf{21,43} & \textbf{21,43} & \textbf{21,43} & 28,57          \\
\#   & 26,19          & 21,43          & 26,19          & \textbf{21,43} \\
*    & 30,95          & 23,81          & \textbf{21,43} & 26,19          \\
@    & 23,81          & 19,05          & 21,43          & \textbf{14,29} \\
\%   & 23,81          & 23,81          & 23,81          & \textbf{21,43} \\
=    & 40,48          & 38,10          & \textbf{28,57} & 30,95          \\
$\epsilon$   & 30,95          & 30,95          & 26,19          & \textbf{23,81} \\
$\alpha$ & 26,19          & 26,19          & \textbf{19,05} & 21,43          \\ \hline
Average & 27.98          & 25.60          & \textbf{23.51} & \textbf{23.51}
\end{tabular}
\end{table}

\begin{table}[h]
\centering
\caption{Performance as EER(\%) of each individual number on the evaluation dataset of e-BioDigitDB.}
\label{exp1_eBioDigit_numbers}
\begin{tabular}{ccccc}
     & DTW   & SW-DTW         & RNNs          & TA-RNNs       \\ \hline
0    & 34.90 & 34,59          & 31,25          & \textbf{29,51} \\
1    & 32.30 & 30,96          & \textbf{27,33} & 28,78          \\
2    & 32,80 & 31,40          & 29,36          & \textbf{25,87} \\
3    & 35.00 & 31,69          & 30,67          & \textbf{26,89} \\
4    & 23.50 & \textbf{18,75} & 25,58          & 21,22          \\
5    & 24.40 & 22,09          & 22,67          & \textbf{19,62} \\
6    & 36,90 & 34,16          & 34,88          & \textbf{26,74} \\
7    & 22,50 & \textbf{20,20} & 25,58          & 23,84          \\
8    & 26,00 & 24,27          & 22,38          & \textbf{19,19} \\
9    & 29.60 & 26,89          & 25,29          & \textbf{21,95} \\ \hline
Average & 29.82 & 27.50          & 27.50          & \textbf{24.36}
\end{tabular}
\end{table}

\subsection{One-Character Analysis}\label{subsec_exp1}
This section analyses the potential of each individual character for the task of user authentication. Four different touch biometric systems are considered in the analysis: \textit{i)} the original DTW system considered in previous studies~\cite{2019_TMC_BioTouchPass_Tolosana,2019_CVPRw_MobileTouch_Tolosana}, \textit{ii)} the recent SW-DTW approach that overcomes the incorrect alignments generated by the original DTW~\cite{folgado2018time}, \textit{iii)} the state-of-the-art RNN Siamese architecture proposed in~\cite{2019_TMC_BioTouchPass_Tolosana}, and finally \textit{iv)} the TA-RNN Siamese architecture proposed in this work that combines SW-DTW alignment with RNNs. For the development of the systems, we follow the experimental protocol described in Sec.~\ref{experimental_protocol}. Samples from all characters of the development datasets of both MobileTouchDB and e-BioDigitDB databases are considered together during training as we intend to distinguish between genuine and impostor handwritten samples regardless of the user and the character. This approach resulted in better generalisation results compared with the case of training one system per user and character. Therefore, both RNNs and TA-RNNs systems are trained considering two different cases: \textit{i)} pairs of genuine characters drawn by the same user, and \textit{ii)} pairs of genuine and impostor characters, one performed by the claimed user and the other one by the impostor. 

Both RNN and TA-RNN systems have been implemented under Keras framework using Tensorflow as back-end, with a NVIDIA GeForce RTX 2080 Ti GPU. The weights of the BLSTM and feed-forward layers are initialised by random values drawn from the zero-mean Gaussian distribution with standard deviation 0.05. Adam optimiser is considered with default parameters (learning rate of 0.001) and a loss function based on binary cross-entropy

Tables~\ref{exp1_mobileTouchDB_letters} and~\ref{exp1_mobileTouchDB_numbers} show the results of each character over the evaluation dataset of MobileTouchDB, grouped according to their corresponding acquisition block. The evaluation results achieved over the e-BioDigitDB database are also depicted in Table~\ref{exp1_eBioDigit_numbers}. Finally, we also provide the average EER obtained for each authentication system and acquisition block for a better comprehension of the results. 

We first compare the system performance results when drawing letters (Table~\ref{exp1_mobileTouchDB_letters}). Analysing upper-case letters, the original DTW system achieves an average 28.57\% EER, being the letter ``X" the one that provides the best system performance with a final 19.05\% EER. The SW-DTW system outperforms the original DTW system, achieving an average 27.06\% EER, which demonstrates the better alignment carried out using this improved DTW version. Both DTW approaches are also compared with state-of-the-art recurrent neural networks. The RNN system outperforms the results achieved using both DTW and SW-DTW, achieving an average 23.72\% EER, an absolute improvement of 4.85\% and 3.34\% EERs compared with DTW and SW-DTW, respectively. Finally, our proposed TA-RNN system is also evaluated in the same conditions, outperforming the results achieved by the other biometric systems for most upper-case characters. Our TA-RNN system obtains an average 20.99\% EER, outperforming in large margin the results achieved by DTW approaches, and also the state-of-the-art RNN system with an absolute improvement of 2.73\% EER. It is interesting to remark the considerable improvements achieved by our proposed TA-RNN system in some upper-case letters such as ``A", ``B", and ``Z", with absolute improvements higher than 4.5\% EER compared with the second best approach. Finally, the proposed TA-RNN system achieves the best system performance result with an EER as low as 11.90\% for the letter ``B", achieving absolute improvements of 7.15\%, 7.15\%, and 4.77\% EERs compared with DTW, SW-DTW, and RNNs, respectively. These results remark the potential of our proposed TA-RNN system, outperforming all previous results achieved in the literature.


We now compare the results achieved when drawing upper- and lower-case letters (Table~\ref{exp1_mobileTouchDB_letters}). In general, better results are obtained when drawing lower-case letters. For the DTW and SW-DTW systems, lower-case letters achieve average absolute improvements of 1.94\% and 2.19\% EERs in comparison with upper-case letters, respectively. Similar improvements are also produced for both RNNs and TA-RNNs, with average absolute improvements of 3.52\% and 2.29\% EERs, respectively. These results remark the higher discriminative power of lower-case letters compared with upper-case letters. We believe this is produced because most upper-case letters are based mostly on simple straight strokes, and not so much in curved strokes, providing therefore less variability among users. In addition, we usually write using lower-case letters, adapting our original writing model to more user-specific features compared with upper-case letters. One example that justifies our hypothesis is letter ``r/R". Letter ``r" provides the best result with a 11.90\% EER for the TA-RNN system. However, the EER increases up to 16.67\% when using letter ``R". Similar observations apply to other letters such as ``v/V" and ``y/Y". Nevertheless, there are some cases where both upper- and lower-case letters obtain very similar results, such as letters ``s/S" and ``g/G" as they are mainly curved strokes in both lower- and upper-case letters. Finally, the same trends along the authentication systems are observed for lower-case letters. Our TA-RNN system obtains an average 18.70\% EER, achieving average absolute improvements of 7.93\%, 6.1\%, and 1.5\% EERs compared with DTW, SW-DTW, and RNNs, respectively.

Characters based on symbols and numbers are also considered in this analysis. Tables~\ref{exp1_mobileTouchDB_numbers} and~\ref{exp1_eBioDigit_numbers} include the results achieved using the evaluation datasets of MobileTouchDB and e-BioDigitDB, respectively. For the case of drawing numbers, our proposed TA-RNN system outperforms the other systems for both MobileTouchDB and e-BioDigit databases. For the MobileTouchDB database (Table~\ref{exp1_mobileTouchDB_numbers}), the proposed TA-RNNs achieves an average 20.24\% EER whereas for the e-BioDigitDB database this result increases up to 24.36\% EER. We believe this is produced due to two main reasons: \textit{i)} the time gap considered between train and test samples in the MobileTouchDB is about two days (i.e., s1 vs. s2) whereas in the e-BioDigitDB this time gap increases up to three weeks; and \textit{ii)} the different acquisition scenario considered, office-like (e-BioDigitDB) and in the wild (MobileTouchDB). Finally, the symbol characters acquired in MobileTouchDB are also analysed in Table~\ref{exp1_mobileTouchDB_numbers}. In general, we can observe the same trends described before. However, it is interesting to remark the lower discriminative power of symbols compared with the other characters of MobileTouchDB when considering our proposed TA-RNN system.

\begin{figure}[t]
\begin{center}
   \includegraphics[width=0.9\linewidth]{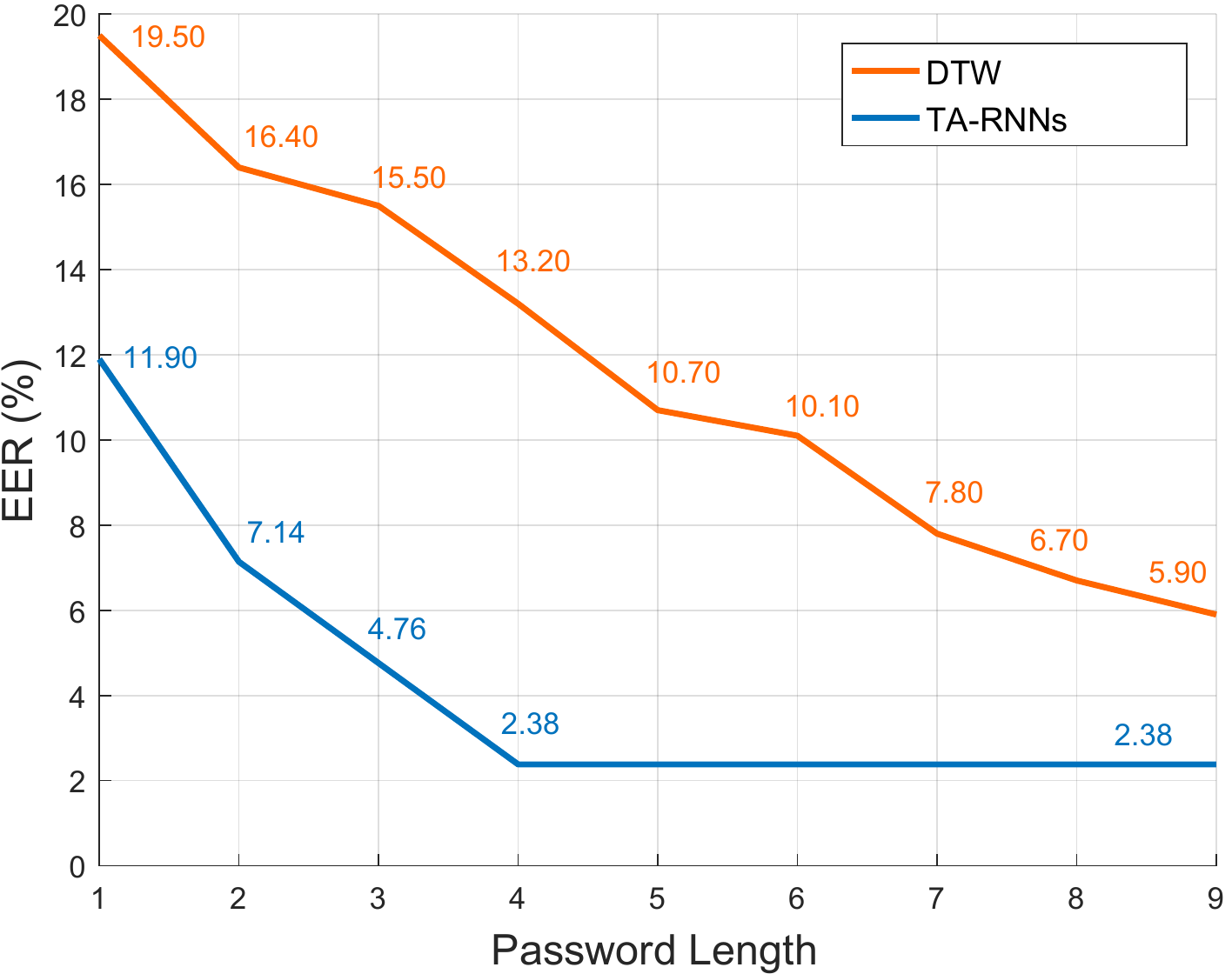}
\end{center}
   \caption{System performance in terms of EER (\%) when increasing the length of the password for the MobileTouchDB evaluation dataset}
\label{exp2_MobileTouchDB}
\end{figure}

\begin{figure*}[h]
  \centering
    \includegraphics[width=0.85\linewidth]{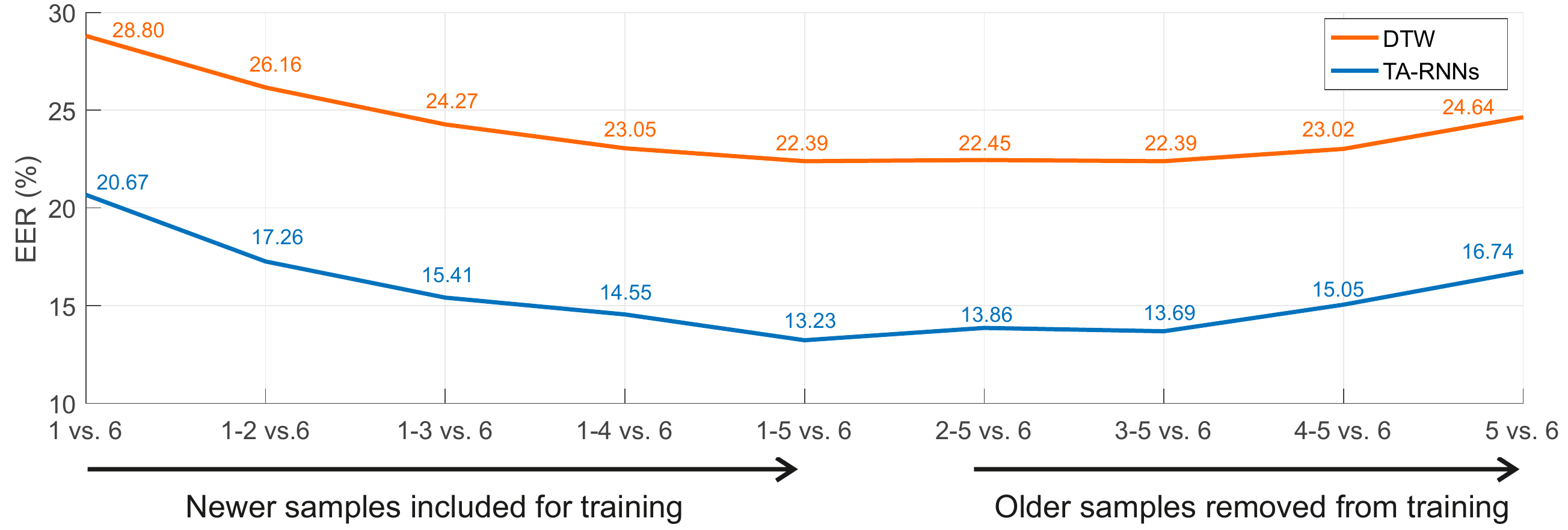}
  \caption{Template update analysis: For each experiment, the first number indicates the training sessions, and the second one the test session (i.e., the sixth session in all experiments). We first add training samples from sessions closer to the test. Then, we remove samples from older sessions to the test.}
  \label{aging_analysis}
\end{figure*}

\subsection{Character Combination Analysis}\label{subsec_exp2}
This section evaluates the robustness of our proposed touchscreen password biometric approach when increasing the length of the password from 1 to 9 characters. The same touch biometric systems studied in Sec.~\ref{subsec_exp1} are considered in this experiment. The final score of each password is produced by fusing the different one by one character score comparisons using the sum of the scores~\cite{2018_INFFUS_MCSreview2_Fierrez}. Fig.~\ref{exp2_MobileTouchDB} shows the evolution of the system performance in terms of EER (\%) when increasing the length of the password for the MobileTouchDB evaluation dataset. Passwords are created following the results extracted in the one-character analysis of Sec.\ref{subsec_exp1}, including the top ranked most discriminative characters in order, one at a time, e.g., the ``B" and ``r" characters are used for a 2-digit password for the TA-RNN system.



Analysing the results obtained in Fig.~\ref{exp2_MobileTouchDB}, our proposed TA-RNN system further outperforms the results achieved in~\cite{2019_CVPRw_MobileTouch_Tolosana} using the original DTW system. For example, for a 3-digit password, our proposed TA-RNN system achieves an absolute improvement of 10.74\% EER compared with the DTW system. In general, we achieve an average absolute improvement of 7.52\% EER when considering all length passwords (from 1 to 9 characters). Our proposed TA-RNN system achieves a final 2.38\% EER using just a 4-digit password.  However, for the DTW system, the best system performance result is a 5.9\% EER when considering a 9-digit password. The results achieved in this experiment prove the success of our proposed touchscreen password biometric approach and also the proposed TA-RNN system in terms of both usability (just a 4-digit password is required to achieve a 2.38\% EER, not 9) and system performance (2.38\% EER vs. 5.90\% EER). Finally, it is important to note that in Fig.~\ref{exp2_MobileTouchDB} no further improvement is observed  by our proposed TA-RNN system when passwords comprise more than 4 characters. This might be produced due to passwords are created including the top ranked most discriminative characters, one at a time, avoiding repetitions. As a result, the discriminative power of the new characters included in the password is not enough to further improve the system performance.



\subsection{Template Update Analysis}\label{subsec_exp3}
This section conducts some experiments in order to analyse the inter-session variability effect on the system performance. MobileTouchDB is considered in this experiment, with samples acquired in 6 different acquisition sessions with different time gaps between them, emulating practical scenarios. First, we focus on how the system performance can be improved along time when using samples from different acquisition sessions as training samples. Fig.~\ref{aging_analysis} shows the average EER (\%) of all 72 individual character comparisons of the evaluation dataset of MobileTouchDB. For each experiment, the first number indicates the training sessions (one sample per session), and the second one indicates the test session (i.e., the sixth session in all experiments). We first add training samples from sessions closer to the test. Then, we remove samples from older sessions in time.

First, we analyse the scenario where samples from sessions closer to the test are included for training. The results achieved in Fig.~\ref{aging_analysis} show very similar trends for both DTW and the proposed TA-RNN systems. The system performance improves when we include samples from different acquisition sessions. For the DTW system, an average 28.80\% EER is achieved when training samples are coming only from the first session (1 vs. 6). This system performance is further improved when increasing the number of training samples, e.g., an absolute improvement of 5.41\% EER is achieved when using five training samples (1-5 vs. 6). Then, when we remove samples from sessions older to the test, the system performance gets worse. For the DTW system, an average 24.64\% EER is achieved when using just one sample from the previous session to the test (5 vs. 6), an average 2.25\% worsening compared with the case of using five training samples (1-5 vs. 6). 

The same trend is observed when using our proposed TA-RNN system, but with a high system performance improvement compared with the DTW system. Our proposed system achieves an average absolute improvement of 8.52\% EER. The best system performance achieved is a final 13.23\% EER, which is 9.16\% EER lower compared with the best result achieved by the DTW system. These results prove the robustness of our proposed TA-RNN approach with time, when samples form multiple acquisition sessions can be considered for training. 

\begin{figure}[t]
  \centering
    \includegraphics[width=0.89\linewidth]{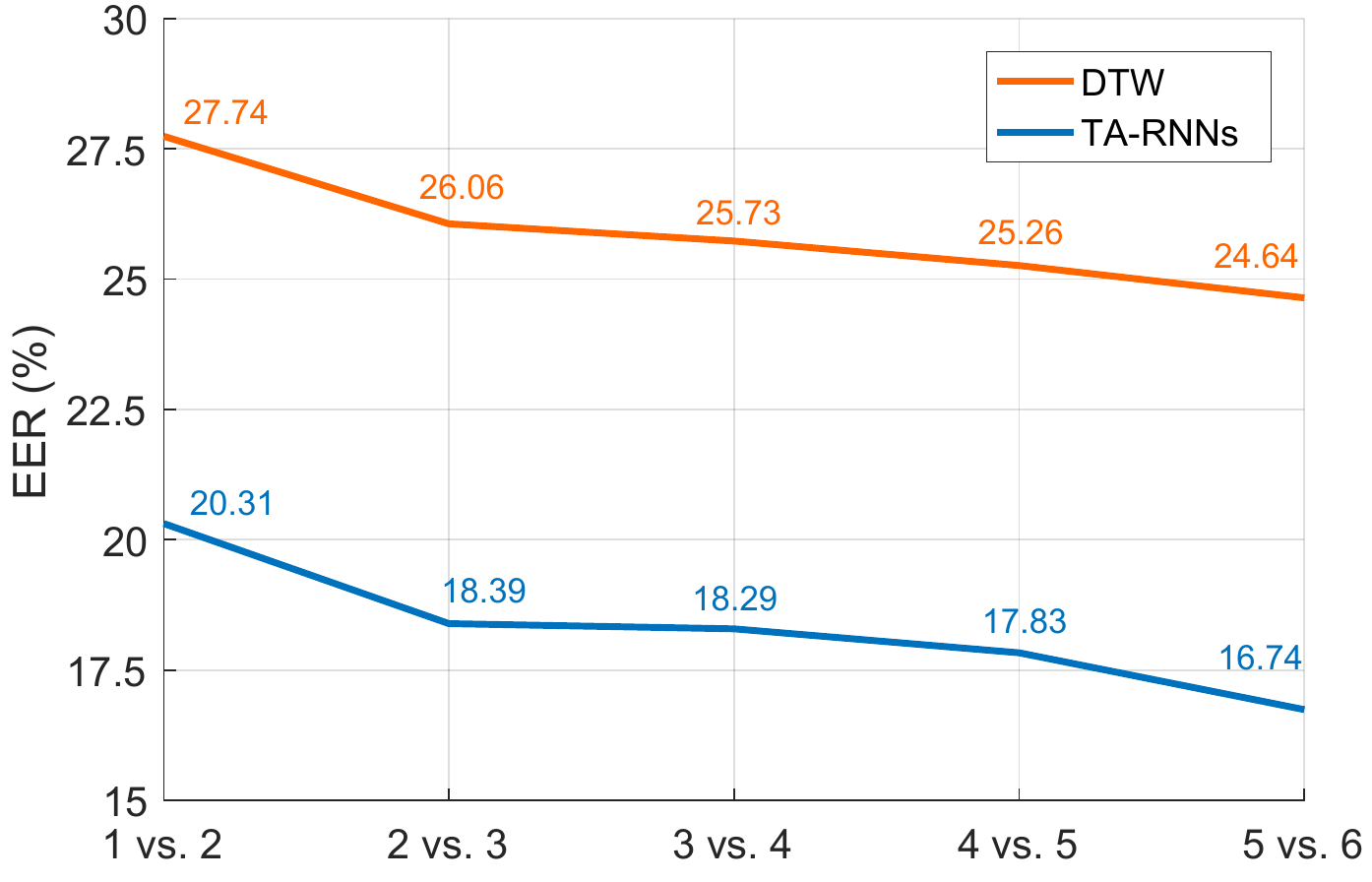}
  \caption{Intra-user analysis on the evaluation dataset of MobileTouchDB.}
  \label{learning_experiment}
\end{figure}

Finally, we also analyse the intra-user variability while drawing the characters along time. Fig.~\ref{learning_experiment} shows the system performance in terms of EER(\%) on the evaluation dataset of MobileTouchDB. Each experiment compares samples from one session with the same samples but for the next session in order to analyse the intra-user variability and the learning curve of the users. For both DTW and TA-RNN systems, the same trend is observed: the system performance improves with time. For the DTW system, an absolute improvement of 3.10\% EER is achieved when we compare the 1 vs. 2 case with the 5 vs. 6 case whereas for the TA-RNN system, an absolute improvement of 3.57\% EER is achieved for the same comparison. These results highlight how users tend to feel more comfortable with the system along time, reducing its variability while drawing the characters, and therefore, being possible to further improve the system performance with the number of acquisition sessions.

\section{TA-RNNs: Further Applications}\label{subsec_exp3}
\begin{figure*}[t]
\centering
\begin{subfigure}[tb]{0.28\textwidth}
\centering
\centerline{\includegraphics[width=\linewidth]{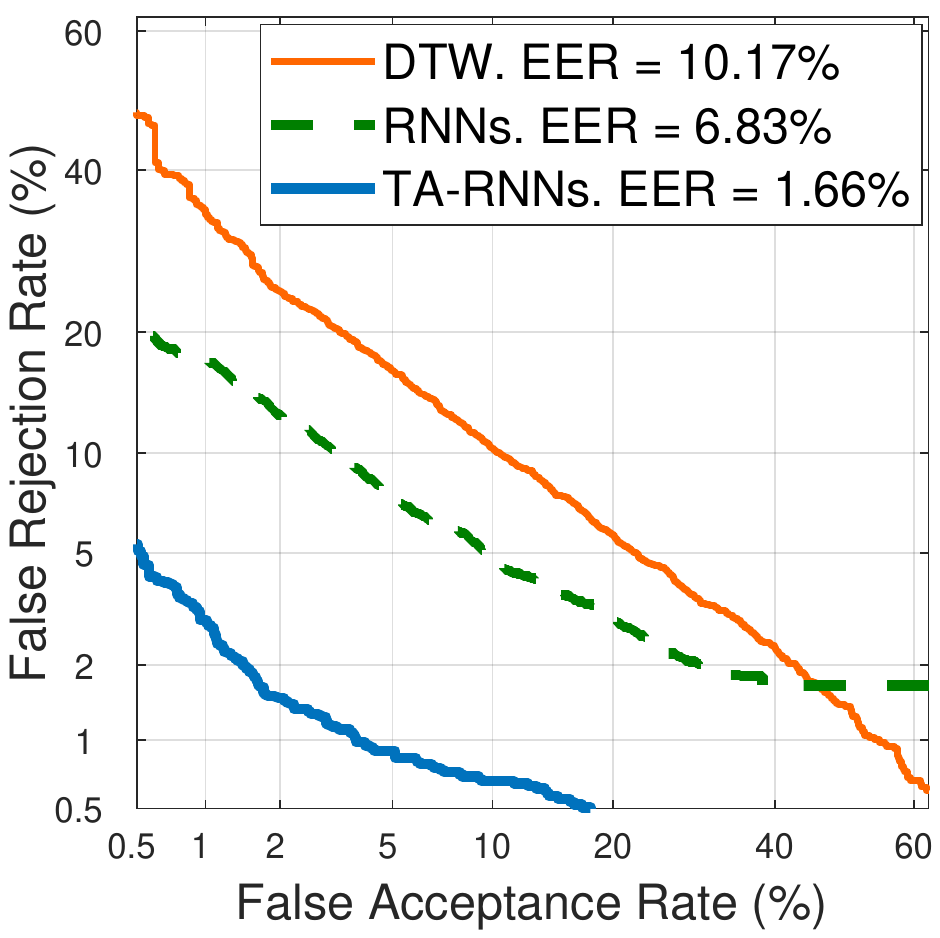}}
\caption{Skilled} \label{fig:DET_stylus}
\end{subfigure}
\begin{subfigure}[tb]{0.28\textwidth}
\centerline{\includegraphics[width=\linewidth]{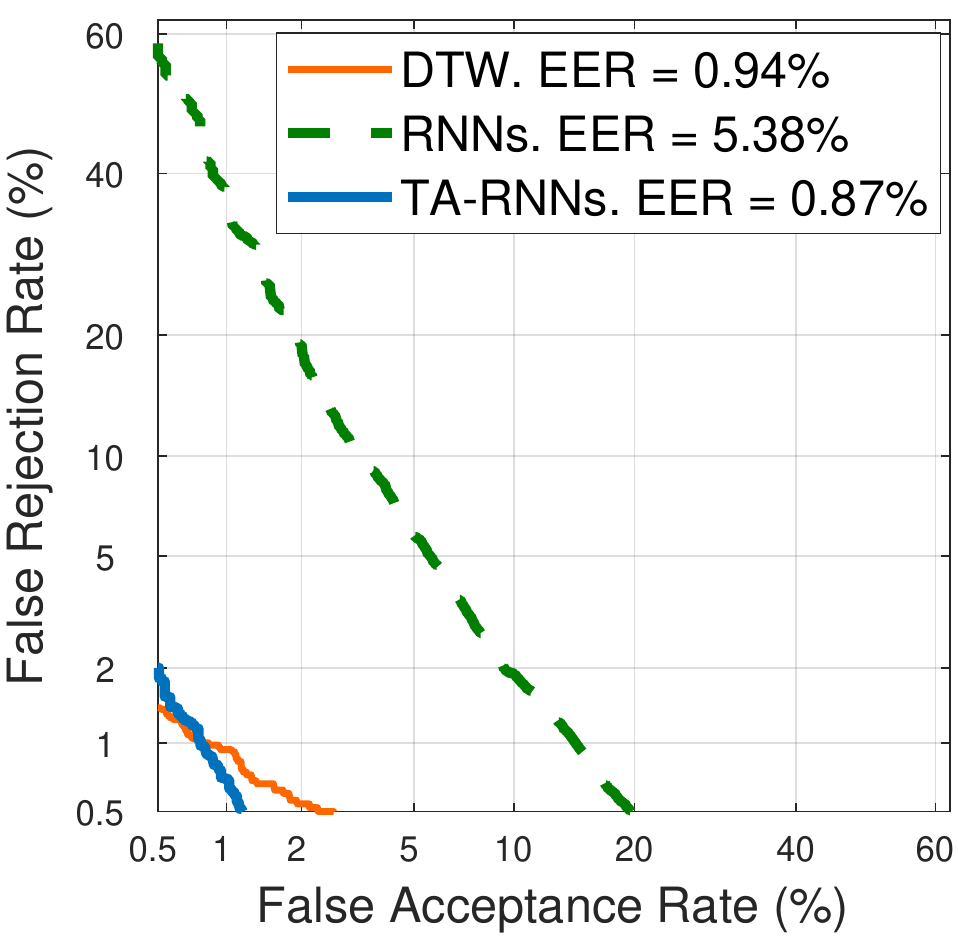}}
\caption{Random} \label{fig:DET_finger}
\end{subfigure}
\caption{System performance results of our proposed TA-RNN approach for the task of on-line signature verification.} \label{fig:DET_signature}
\end{figure*}

Finally, our proposed TA-RNN system can be applied to many other different applications in which a direct comparison or verification of time sequence signals is performed, for example, in speech or gait~\cite{graves2013speech, costilla2018analysis}. A related topic is signature verification. In this section we test our proposed TA-RNN approach for on-line signature verification considering the same experimental framework considered in~\cite{2018_IEEEAccess_RNN_Tolosana}, i.e., the first 300 users of BiosecurID are considered for development whereas the remaining 100 users are considered for evaluation. It is important to remark that we consider just one training signature per user in this experiment (i.e., 1vs1 case) to show the real potential of TA-RNNs. Table~\ref{exp_signature_verification} shows the evaluation results in terms of EER (\%) for both skilled and random forgery scenarios~\cite{2018_HanbookBioAntiSpoofing_signature_Tolosana}. We also depict in Fig.~\ref{fig:DET_signature} the results achieved in terms of Detection Error Tradeoff (DET) curves, for completeness. Our proposed TA-RNN system is compared with the state-of-the-art DTW and RNN configurations proposed for signature verification~\cite{2019_ICDAR_DeepSignDB_Tolosana}. For skilled forgery scenarios, our proposed TA-RNN system achieves 1.66\% EER, an absolute improvement of 8.51\% and 5.17\% EERs compared with the DTW and RNN systems respectively, which is a very important reduction of EER. Analysing random forgery scenarios, our proposed TA-RNN system achieves 0.87\% EER, an absolute improvement of 4.51\% EER compared with the RNN system proposed in~\cite{2018_IEEEAccess_RNN_Tolosana}. It is important to remark that in~\cite{2018_IEEEAccess_RNN_Tolosana}, the RNN system was not able to outperform the DTW system for both skilled and random forgery scenarios. Nevertheless, our proposed TA-RNN system can learn better feature representations of both type of impostors after the pre-warping of the time functions, outperforming the state of the art for both types of impostors.

\begin{table}[t]
 \centering
 \caption{Further applications of our proposed TA-RNN system: performance as EER(\%) for signature verification.}
\label{exp_signature_verification}
\begin{tabular}{c|ccc} 
        & DTW   & RNNs & \textbf{TA-RNNs} \\ \hline
Skilled & 10.17 & 6.83  & \textbf{1.66}     \\
Random  & 0.94  & 5.38  & \textbf{0.87}    
\end{tabular}
\end{table}

\section{Conclusions}\label{conclusions}
This work enhances password scenarios through two-factor authentication approaches asking the users to draw each character of the password instead of typing them as usual. The main contributions of this study are as follows: \textit{i)} We present the novel MobileTouchDB public database, acquired in an unsupervised mobile scenario with no restrictions in terms of position, posture, and devices. This database contains more than 64K on-line character samples performed by 217 users, with 94 different smartphone models, and up to 6  acquisition sessions. \textit{ii)} We perform a complete analysis of the proposed approach considering both traditional authentication systems such as DTW and novel approaches based on RNNs. In addition, we present a novel approach named Time-Aligned Recurrent Neural Networks (TA-RNNs). This approach combines the potential of DTW and RNNs to train more robust systems against attacks. 

In this study we have performed a complete analysis of the proposed approach using both MobileTouchDB and e-BioDigitDB. Our proposed TA-RNN system has outperformed the state of the art, achieving a final 2.38\% EER, using just a 4-digit password and one training sample per character. These results encourage the deployment of our approach in comparison with traditional systems where the attack would have 100\% success rate under the same impostor scenario.

In addition, we have also demonstrated the application of our proposed TA-RNNs for another time sequence recognition task, i.e., on-line handwritten signature verification. The proposed TA-RNN system has achieved 1.66\% and 0.87\% EERs for skilled and random forgery scenarios respectively, outperforming in large margin the state of the art.

Future work will be oriented to study the discriminative power of new features acquired in MobileTouchDB such as the area covered by the finger, accelerometer, and gyroscope in order to further improve the system performance~\cite{2019_Passwords_Ross}. We will also analyse the user experience in different acquisition settings through the analysis of the information acquired from Block5 to Block8 of the MobileTouchDB. Finally, we will evaluate the usability and performance improvement of our proposed TA-RNN approach for other behavioural biometric traits such as keystroke biometrics~\cite{2016_IEEEAccess_KBOC_Aythami} and also on identification scenarios~\cite{zhang2016end}.

\section*{Acknowledgments}
\small{This work has been supported by projects: BIBECA (RTI2018-101248-B-I00 MINECO/FEDER), Bio-Guard (Ayudas Fundaci\'on BBVA a Equipos de Investigaci\'on Cient\'ifica 2017) and by UAM-CecaBank. Spanish Patent Application (P202030060).}



%

%

{
\bibliographystyle{IEEEtran}
\bibliography{egbib2}
}

\begin{IEEEbiography}[{\includegraphics[width=1in,height=1.25in,clip,keepaspectratio]{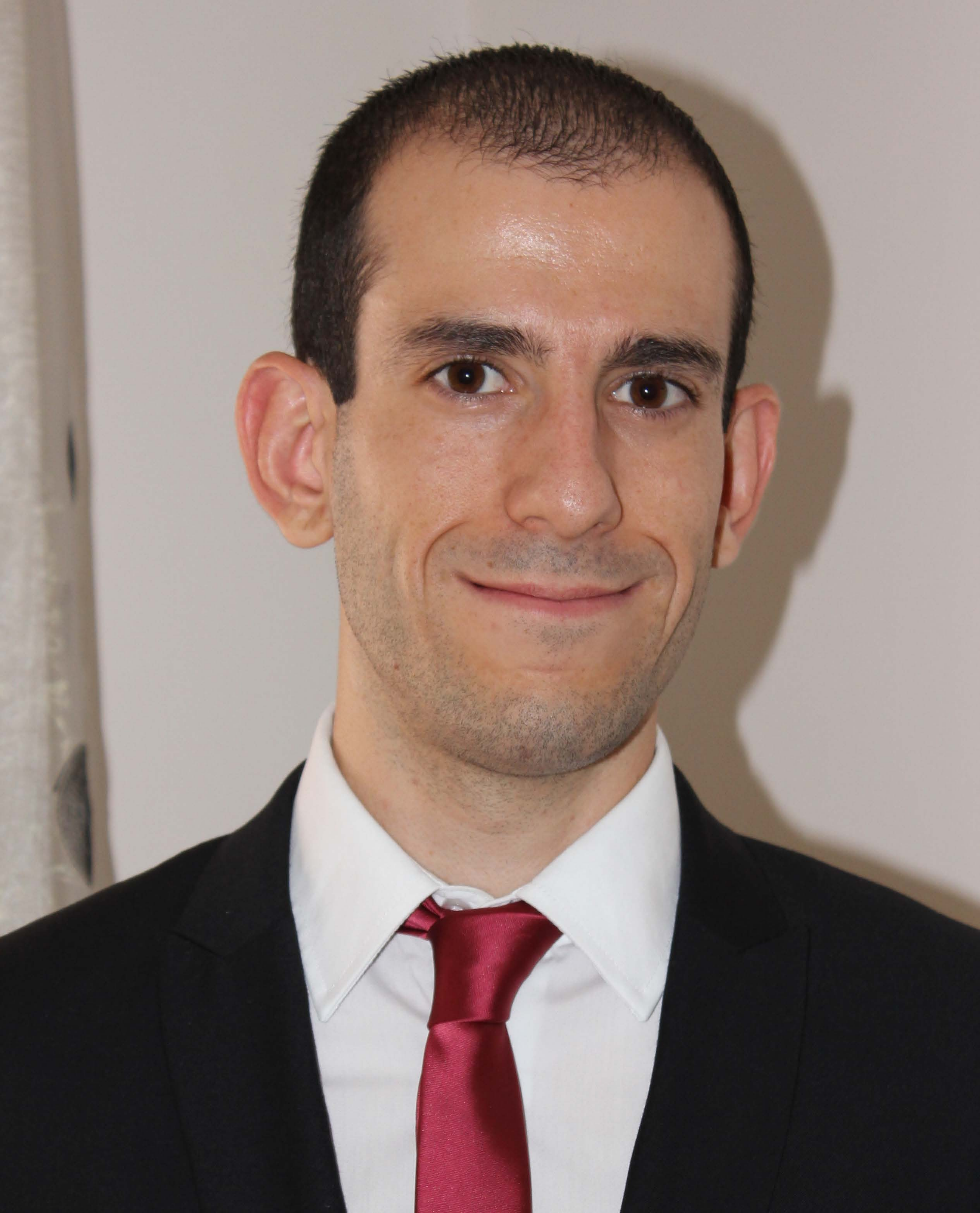}}]%
{Ruben Tolosana}
received the M.Sc. degree in Telecommunication Engineering, and his Ph.D. degree in Computer and Telecommunication Engineering, from Universidad Autonoma de Madrid, in 2014 and 2019, respectively. In April 2014, he joined the Biometrics and Data Pattern Analytics - BiDA Lab at the Universidad Autonoma de Madrid, where he is currently collaborating as a PostDoctoral researcher. Since then, Ruben has been granted with several awards such as the FPU research fellowship from Spanish MECD (2015), and the European Biometrics Industry Award (2018). His research interests are mainly focused on signal and image processing, pattern recognition, and machine learning, particularly in the areas of face manipulation, human-computer interaction and biometrics. He is author of several publications and also collaborates as a reviewer in many different high-impact conferences (e.g., ICDAR, IJCB, ICB, BTAS, EUSIPCO, etc.) and journals (e.g., IEEE TPAMI, TCYB, TIFS, TIP, ACM CSUR, etc.). Finally, he has participated in several National and European projects focused on the deployment of biometric security through the world.
\end{IEEEbiography}

\begin{IEEEbiography}[{\includegraphics[width=1in,height=1.25in,clip,keepaspectratio]{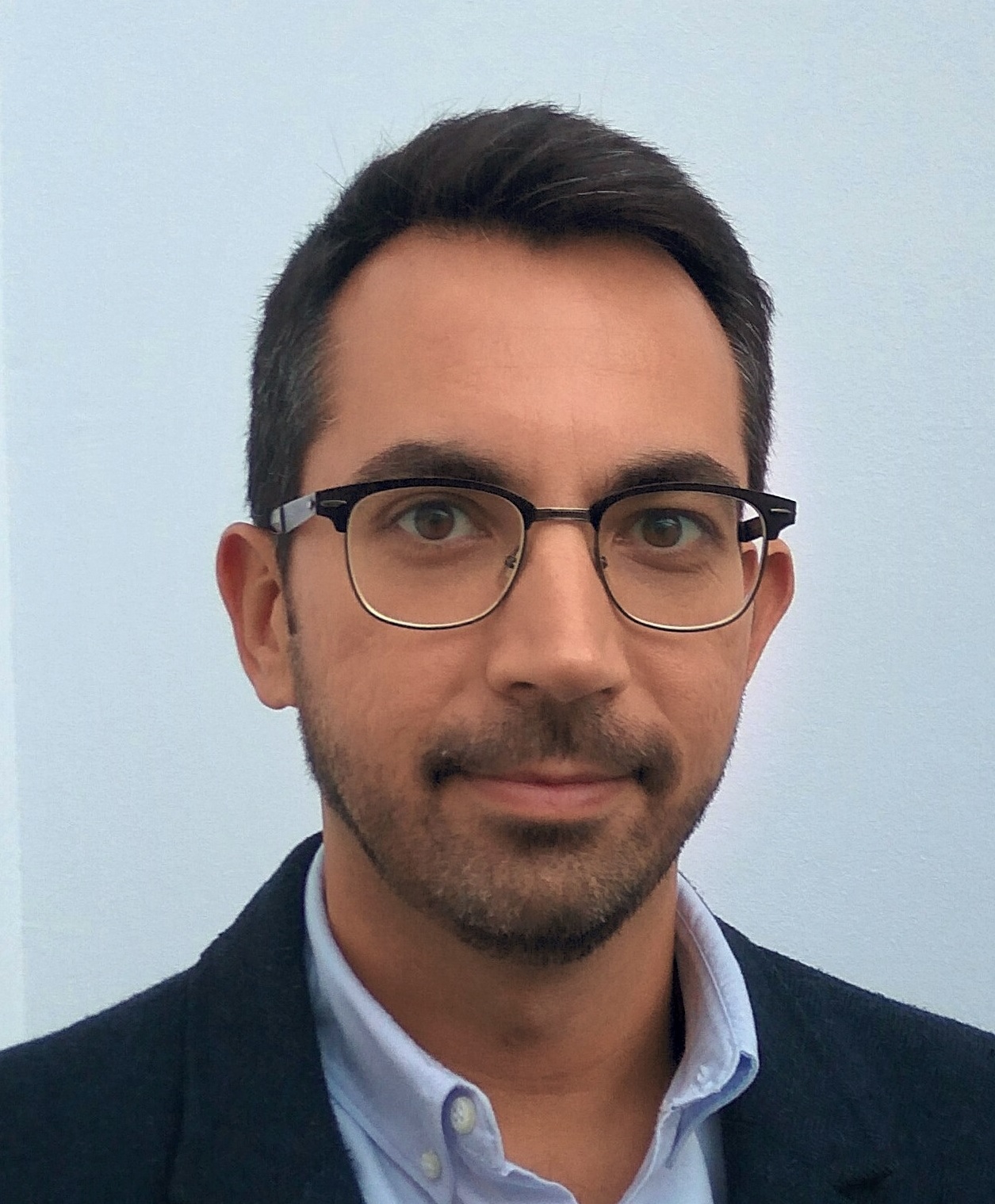}}]{Ruben Vera-Rodriguez} received the M.Sc. degree in telecommunications engineering from Universidad de Sevilla, Spain, in 2006, and the Ph.D. degree in electrical and electronic engineering from Swansea University, U.K., in 2010. Since 2010, he has been affiliated with the Biometric Recognition Group, Universidad Autonoma de Madrid, Spain, where he is currently an Associate Professor since 2018. His research interests include signal and image processing, pattern recognition, and biometrics, with emphasis on signature, face, gait verification and forensic applications of biometrics. He is actively involved in several National and European projects focused on biometrics. Ruben has been Program Chair for the IEEE 51st International Carnahan Conference on Security and Technology (ICCST) in 2017; and the 23rd Iberoamerican Congress on Pattern Recognition (CIARP 2018) in 2018.
\end{IEEEbiography}

\begin{IEEEbiography}[{\includegraphics[width=1in,height=1.25in,clip,keepaspectratio]{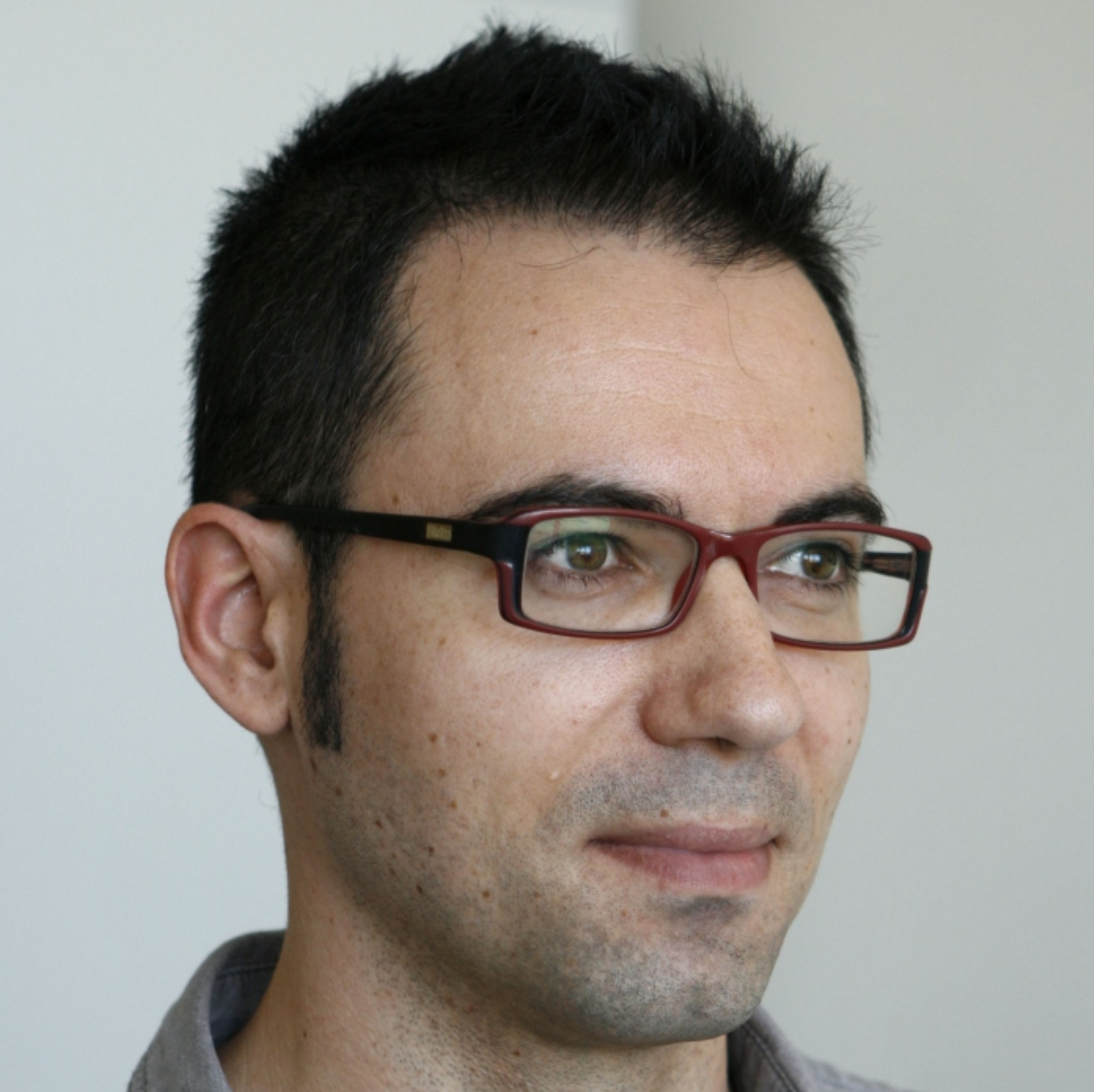}}]{Julian Fierrez} received the M.Sc. and Ph.D. degrees in telecommunications engineering from the Universidad Politecnica de Madrid, Spain, in 2001 and 2006, respectively. Since 2002, he has been with the Biometric Recognition Group, Universidad Politecnica de Madrid. Since 2004, he has been with the Universidad Autonoma de Madrid, where he is currently an Associate Professor. From 2007 to 2009, he was a Visiting Researcher with Michigan State University, USA, under a Marie Curie Fellowship. His research interests include signal and image processing, pattern recognition, and biometrics, with an emphasis on multibiometrics, biometric evaluation, system security, forensics, and mobile applications of biometrics. He has been actively involved in multiple EU projects focused on biometrics (e.g., TABULA RASA and BEAT), and has attracted notable impact for his research. He was a recipient of a number of distinctions, including the EAB European Biometric Industry Award 2006, the EURASIP Best Ph.D. Award 2012, the Miguel Catalan Award to the Best Researcher under 40 in the Community of Madrid in the general area of science and technology, and the 2017 IAPR Young Biometrics Investigator Award. He is an Associate Editor of the IEEE TRANSACTIONS ON INFORMATION FORENSICS AND SECURITY and the IEEE TRANSACTIONS ON IMAGE PROCESSING.
\end{IEEEbiography}

\begin{IEEEbiography}[{\includegraphics[width=1in,height=1.25in,clip,keepaspectratio]{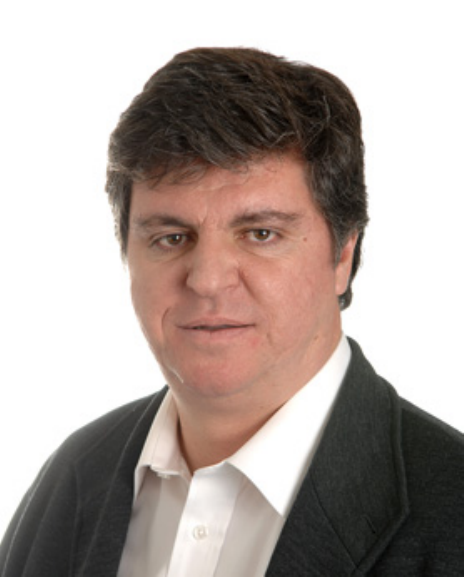}}]%
{Javier Ortega-Garcia}
received the M.Sc. degree in electrical engineering and the Ph.D. degree (cum laude) in electrical engineering from Universidad Politecnica de Madrid, Spain, in 1989 and 1996, respectively. He is currently a Full Professor at the Signal Processing Chair in Universidad Autonoma de Madrid - Spain, where he holds courses on biometric recognition and digital signal processing. He is a founder and Director of the BiDA-Lab, Biometrics and Data Pattern Analytics Group. He has authored over 300 international contributions, including book chapters, refereed journal, and conference papers. His research interests are focused on biometric pattern recognition (on-line signature verification, speaker recognition, human-device interaction) for security, e-health and user profiling applications. He chaired Odyssey-04, The Speaker Recognition Workshop, ICB-2013, the 6th IAPR International Conference on Biometrics, and ICCST2017, the 51st IEEE International Carnahan Conference on Security Technology.
\end{IEEEbiography}

\end{document}